\documentclass[10pt,twocolumn,letterpaper]{article}
\usepackage{iccv}
\usepackage{times}
\usepackage{epsfig}
\usepackage{graphicx}
\usepackage{amsmath}
\usepackage{amssymb}

\graphicspath{{figures/}}

\usepackage[utf8]{inputenc} 
\usepackage[T1]{fontenc}    
\usepackage{url}            
\usepackage{booktabs}       
\usepackage{amsfonts}       
\usepackage{nicefrac}       
\usepackage{microtype}      
\usepackage{algorithm}
\usepackage{algorithmic}
\usepackage{array}
\usepackage{bm}
\usepackage{multirow}
\usepackage{color}
\usepackage{bm}
\usepackage{fancyhdr}

\usepackage[pagebackref=true,breaklinks=true,colorlinks,bookmarks=false]{hyperref}

\iccvfinalcopy 

\def\iccvPaperID{****} 

\ificcvfinal\fi

\begin{document}
	
	\title{FREE: Feature Refinement for Generalized Zero-Shot Learning}
	
	
	\author{Shiming Chen$^{1}$, Wenjie Wang$^{1}$, Beihao Xia$^{1}$, Qinmu Peng$^{1}$,  Xinge You$^{1\thanks{Corresponding author}}$, Feng Zheng$^{2}$, Ling Shao$^{3}$\\
		$^{1}$Huazhong University of Science and Technology (HUST), China \\
		$^{2}$Southern University of Science and Technology (SUSTech), China\\
		$^{3}$Inception Institute of Artificial Intelligence (IIAI), UAE\\
		{\tt\small \{shimingchen,pengqinmu,youxg\}@hust.edu.cn \quad zfeng02@gmail.com \quad	ling.shao@ieee.org}}
	
	\maketitle
	
	\thispagestyle{fancy}	
	\fancyhf{ } 
	\lhead{\parbox{18cm}{\centering 2021 IEEE/CVF International Conference on Computer Vision (ICCV)}}
	\renewcommand{\headrulewidth}{0pt}

	
	\begin{abstract}
		Generalized zero-shot learning (GZSL) has achieved significant progress, with many efforts dedicated to overcoming the problems of visual-semantic domain gap and seen-unseen bias. However, most existing methods directly use feature extraction models trained on ImageNet alone, ignoring the cross-dataset bias between ImageNet and GZSL benchmarks. Such a bias inevitably results in poor-quality visual features for GZSL tasks, which potentially limits the recognition performance on both seen and unseen classes. In this paper, we propose a simple yet effective GZSL method, termed feature refinement for generalized zero-shot learning (FREE), to tackle the above problem. FREE employs a feature refinement (FR) module that incorporates \textit{semantic$\rightarrow$visual} mapping into a unified generative model to refine the visual features of seen and unseen class samples. Furthermore, we propose a self-adaptive margin center loss (SAMC-loss) that cooperates with a semantic cycle-consistency loss to guide FR to learn class- and semantically-relevant representations, and concatenate the features in FR to extract the fully refined features. Extensive experiments on five benchmark datasets demonstrate the significant performance gain of FREE over its baseline and current state-of-the-art methods. Our codes are available at \url{https://github.com/shiming-chen/FREE} . 
	\end{abstract}
	
	\section{Introduction}\label{sec1}
	
	A key challenge of artificial intelligence is to generalize machine learning models from seen data to unseen scenarios. Zero-shot learning (ZSL) is a typical research topic targeting this goal \cite{Lampert2009LearningTD,Larochelle2008ZerodataLO,Palatucci2009ZeroshotLW}. ZSL aims to classify the images of unseen classes by constructing a mapping relationship between the semantic and visual domains. It is usually based on the assumption that both seen and unseen classes can be described through a set of semantic vectors, e.g., sentence embeddings \cite{Reed2016LearningDR}, and attribute vectors \cite{Lampert2014AttributeBasedCF}, in the same semantic space. According to their classification range, ZSL methods can be categorized into conventional ZSL (CZSL) and generalized ZSL (GZSL) \cite{Xian2019ZeroShotLC}. CZSL aims to predict unseen classes, while GZSL can predict both seen and unseen classes. Recently, GZSL has attracted more attention as it is more realistic and challenging. We are thus also interested in the GZSL setting in this paper.

	\begin{figure}[t]
		\begin{center}
			\includegraphics[width=8.7cm,height=5.7cm]{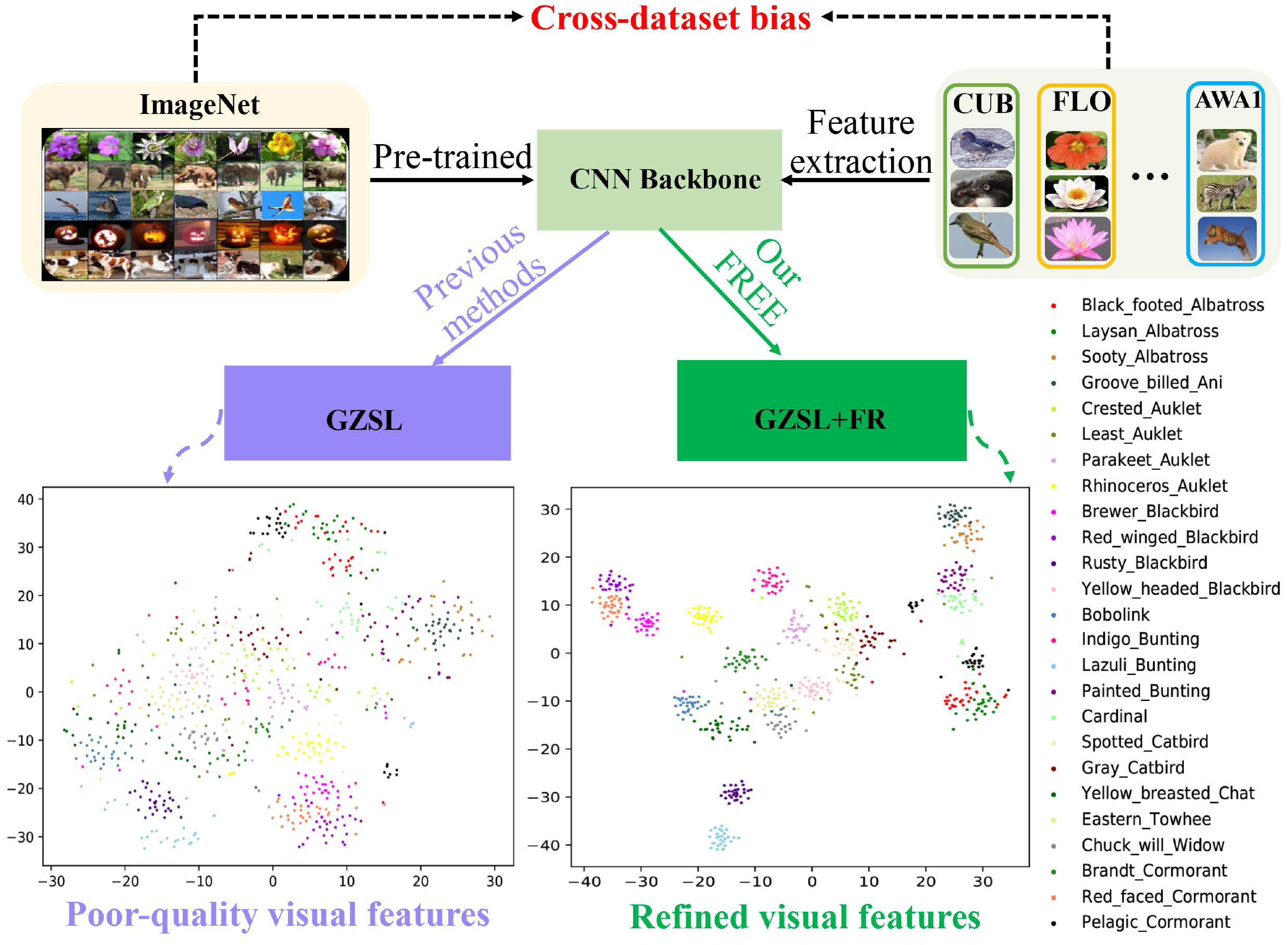}\hspace{-4mm}
			\vspace{-4mm}
			\caption{The core idea of our FREE. The cross-dataset bias between ImageNet and GZSL benchmarks (e.g., CUB) is harmful for feature extraction from GZSL benchmarks, which results in poor-quality visual features for unsatisfying performance in GZSL. Our FREE refines the visual features and improves the \textit{semantic$\rightarrow$visual} mapping using feature refinement (FR) in a unified network for GZSL classification.}
			\label{fig:problem-definition}
		\end{center}
	\end{figure}
	
	GZSL has achieved significant progress, with many efforts focused on the problems of \textit{visual-semantic domain gaps} \cite{Lampert2014AttributeBasedCF,Akata2016LabelEmbeddingFI,Akata2015EvaluationOO,Wang2017ZeroShotVR,Vyas2020LeveragingSA,Xie2020RegionGE} and \textit{seen-unseen bias} \cite{Xian2018FeatureGN,Mishra2018AGM,Zhang2019CoRepresentationNF,Yu2019ZeroshotLV,Shen2020InvertibleZR,Narayan2020LatentEF,Min2020DomainAwareVB,Huynh2020FineGrainedGZ}. Semantic embedding  \cite{Liu2018GeneralizedZL,Chen2018ZeroShotVR,Li2019RethinkingZL,Zhang2019CoRepresentationNF,Liu2019GraphAA} or generative methods (e.g., variational autoencoders (VAEs) \cite{Arora2018GeneralizedZL,Schnfeld2019GeneralizedZA}, generative adversarial nets (GANs) \cite{Xian2018FeatureGN,Li2019RethinkingZL,Xie2019AttentiveRE,Yu2020EpisodeBasedPG,Keshari2020GeneralizedZL,Vyas2020LeveragingSA}, and generative flows \cite{Shen2020InvertibleZR}) are typically applied to mitigate these challenges. 
	
	An important observation of ours is that the unsatisfying performance in GZSL that still nevertheless exists is closely related to the cross-dataset bias \cite{2011Unbiased}. GZSL models usually extract visual features from coarse- and fine-grained benchmarks (e.g., AWA1 \cite{Lampert2014AttributeBasedCF} and CUB \cite{Welinder2010CaltechUCSDB2}) using a convolutional neural network (CNN) backbone (e.g., ResNet-101 \cite{He2016DeepRL}) pre-trained on ImageNet \cite{Xian2019ZeroShotLC}. However, cross-dataset bias, where the data collection procedure can be biased by human or systematic factors, can lead to a distribution mismatch between two datasets, e.g., Auklets are found in the CUB dataset but not in ImageNet. Thus, it is unwise to directly transfer knowledge from ImageNet to a new dataset for GZSL without any further sequential learning, because cross-dataset bias limits knowledge transfer and results in the extraction of poor-quality visual features from GZSL benchmarks, as shown in Fig. \ref{fig:problem-definition}. Further, the larger the bias between ImageNet and the GZSL benchmark, the poorer the knowledge transfer and feature extraction. Since there is a more obvious bias for fine-grained datasets (e.g., CUB), these typically yield inferior performance to coarse-grained datasets (e.g., AWA) for all GZSL methods. The negative effect of cross-dataset bias on the performance of GZSL has been further validated experimentally. In \cite{Xian2019FVAEGAND2AF}, Xian fine-tuned a ResNet pre-trained on ImageNet using seen classes from GZSL benchmarks. Before fine-tuning, f-VAEGAN achieved a harmonic mean of 64.6\% and 63.5\% on FLO and AWA2, respectively, while these numbers increased to 75.1\% and 65.2\% afterward, as shown in Table \ref{Table:fr_fine}. However, Xian did not analyze or discuss this phenomenon. Further, although fine-tuning may alleviate the cross-dataset bias to a certain degree, it inevitably results in other severer problems, e.g., overfitting \cite{Hendrycks2019UsingPC,Li2020RethinkingTH}. Thus, properly addressing the problem of cross-dataset bias in GZSL has become very necessary. To the best of our knowledge, we are the first to identify this as an open issue in GZSL, which will be tackled in this paper.

	To address the above challenges, we propose a novel GZSL method, termed feature refinement for generalized zero-shot learning (FREE), to further boost the performance of GZSL. 
	In essence, FREE refines visual features in a unified generative model, which simultaneously benefits \textit{semantic$\rightarrow$visual} learning, feature synthesis, and classification. 
	Specifically, we take f-VAEGAN \cite{Xian2019FVAEGAND2AF} as a baseline to learn a \textit{semantic$\rightarrow$visual} mapping.
	To improve the visual features of seen and unseen class samples, we employ a  feature refinement (FR) module, which can be jointly optimized with f-VAEGAN to effectively avoid the drawbacks of fine-tuning. 
	Since class label information is available, we introduce a self-adaptive margin center loss (SAMC-loss) to explicitly encourage intra-class compactness and inter-class separability that can adapt to different datasets, i.e., coarse-grained and fine-grained, and guide FR to learn discriminative class-relevant features.
	Thus, the distributions of different classes can be easily separated, as shown in Fig. \ref{fig:problem-definition}. 
	To better learn semantically-relevant and more discriminative visual features, a semantic cycle-consistency loss is also added after the restitution of features. 
	From the residual information \cite{He2016DeepRL}, we further concatenate the discriminative features of various layers in FR to extract the fully refined features. 
	
	To summarize, this paper provides the following important contributions: 
	\textbf{(1)} We propose a novel GZSL method, termed feature refinement for generalized zero-shot learning (FREE), to address the problem of cross-dataset bias, which can further boost the performance of GZSL. To achieve this goal, a feature refinement (FR) module that cooperates with \textit{semantic$\rightarrow$visual} mapping in a unified framework is explored. Importantly, the two modules can be jointly optimized. 
	\textbf{(2)} We propose a self-adaptive margin center loss (SAMC-loss) to explicitly encourage intra-class compactness and inter-class separability. The SAMC-loss also cooperates with a semantic cycle-consistency constraint to enable FR to learn more discriminative class- and semantically-relevant representations, which are especially important for GZSL. 
	\textbf{(3)}  Extensive experimental results on five benchmarks, i.e., CUB, SUN, FLO, AWA1, and AWA2, clearly demonstrate the advantages of the proposed FREE over its baseline and current state-of-the-art methods.
	
	\begin{figure*}[t]
		\begin{center}
			\includegraphics[width=1\linewidth]{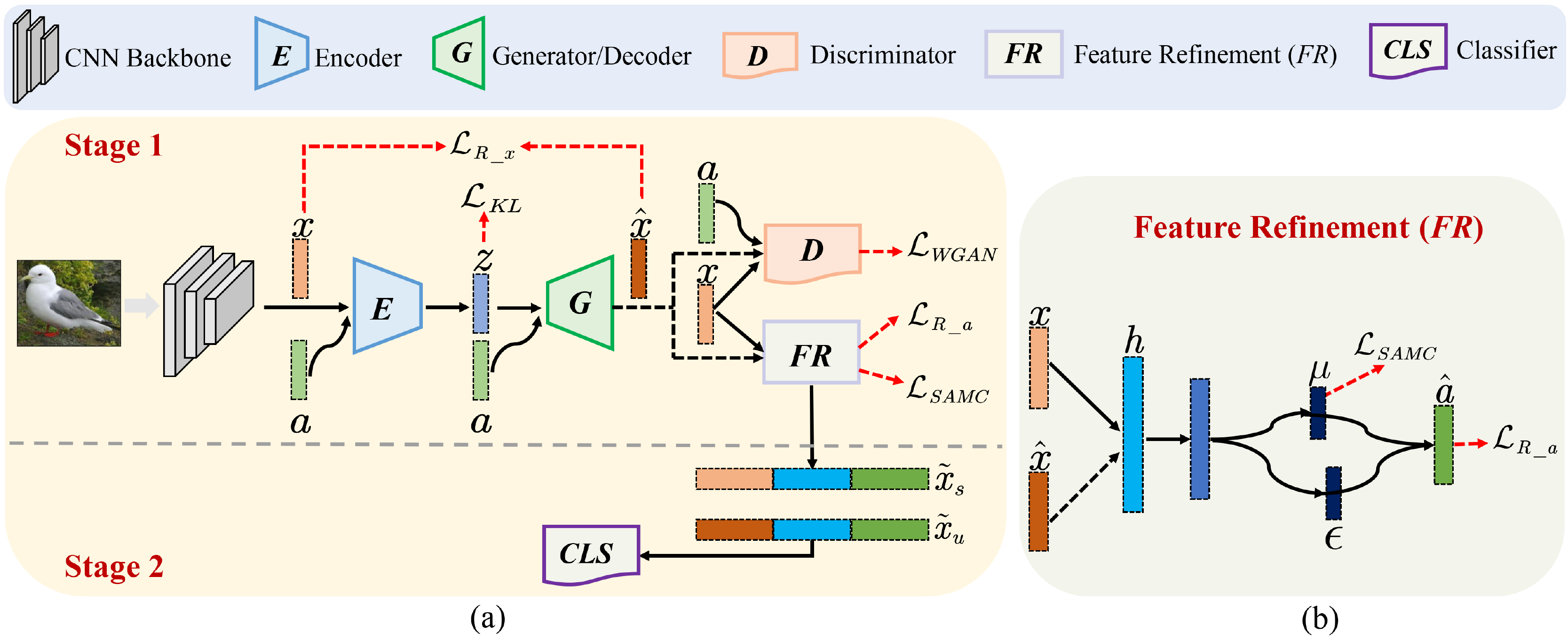}
			\\
			
			\caption{A schematic overview of FREE. (a) FREE consists of a feature generating VAEGAN (f-VAEGAN), a feature refinement (FR) module, and a classifier. In Stage 1, f-VAEGAN aims to learn the \textit{semantic$\rightarrow$visual} mapping ($G$) for visual feature generation, while FR tries to learn discriminative representations for the real/synthesized seen visual features. The two are jointly optimized. In Stage 2, we refine the real seen visual features and the real/synthesized unseen visual features with the trained FR, and they are then fed into a classifier for classification. (b) Proposed FR module. FR learns discriminative features utilizing the SAMC-loss and semantic cycle-consistency loss, and its discriminative features in various layers are then concatenated to extract the fully refined features.} 
			\label{fig:FREE}
		\end{center}
	\end{figure*}

	\section{Related Work}\label{sec2}
	
	\noindent \textbf{Visual-Semantic Domain Gap.} The required knowledge transfer from seen to unseen classes for GZSL relies on semantic embedding. One key task is to bridge the visual and semantic domains \cite{Lampert2014AttributeBasedCF,Akata2016LabelEmbeddingFI,Akata2015EvaluationOO,Wang2017ZeroShotVR,Felix2018MultimodalCG, Vyas2020LeveragingSA}. Since visual features in various forms may convey the same concept, the distribution of instances in the visual space is often distinct from that of their underlying semantics in the semantic space. Thus, there is typically a gap between the two domains, known as the visual-semantic domain gap problem of GZSL \cite{Wang2017ZeroShotVR}. Common space learning  \cite{Changpinyo2017PredictingVE,Zhang2016ZeroShotLV,Xian2016LatentEF,Wang2017ZeroShotVR,Han2020LearningTR}, model parameter transfer \cite{Changpinyo2016SynthesizedCF,Gan2015ExploringSI,Jiang2019TransferableCN} and direct mapping  \cite{Akata2016LabelEmbeddingFI,Akata2015EvaluationOO,Felix2018MultimodalCG,Xian2018FeatureGN,Xian2019FVAEGAND2AF,Narayan2020LatentEF,Huang2019GenerativeDA,Li2017ZeroShotRU,Chen2018ZeroShotVR,Frome2013DeViSEAD} are often used to bridge this gap. Early  mapping works \cite{Akata2016LabelEmbeddingFI,Akata2015EvaluationOO} were carried out by either a classifier or a regression model, depending on the  semantic representation adopted. Most recent mapping works, in contrast, are based on generative models (e.g., VAEs \cite{Arora2018GeneralizedZL,Schnfeld2019GeneralizedZA,Liu2019GraphAA}, GANs \cite{Xian2018FeatureGN,Li2019RethinkingZL,Li2019LeveragingTI,Paul2019SemanticallyAB,Sariyildiz2019GradientMG,Yu2020EpisodeBasedPG}, and generative flows \cite{Shen2020InvertibleZR}). These models not only  learn visual-semantic mapping, but also  generate a great number of feature samples of unseen classes for data augmentation.

	\noindent \textbf{Seen-Unseen Bias.} GZSL methods inevitably encounter a seen-unseen bias problem \cite{Schnfeld2019GeneralizedZA,Xian2019FVAEGAND2AF,Shen2020InvertibleZR,Shen2020InvertibleZR,Liu2018GeneralizedZL,Zhang2019CoRepresentationNF,Ding2019MarginalizedLS,Wu2020SelfSupervisedDG,Jiang2019TransferableCN,Li2019CompressingUI,Atzmon2019AdaptiveCS,Xie2021generalized,Xie2021vman}.  As only seen data is involved during training, most generative GZSL models tend to overfit to seen classes \cite{Liu2018GeneralizedZL,Chen2020ABB,Min2020DomainAwareVB}, i.e., the unseen data generated tends to have the same distribution as the seen categories.  This heavily hampers the classification performance of unseen classes. In \cite{Liu2018GeneralizedZL}, Liu proposed the Deep Calibration Network (DCN) to enable simultaneous calibration of deep networks on the confidence of source classes and the uncertainty of target classes. Zhang \textit{et al.} \cite{Zhang2019CoRepresentationNF} employed a co-representation network to learn a more uniform visual embedding space, effectively bypassing the bias problems and improving classification. Maximum mean discrepancy (MMD) based methods optimize the distribution between real seen features and synthesized unseen features to tackle the bias problem explicitly \cite{Bucher2017GeneratingVR,Shen2020InvertibleZR}. 
	
	\noindent \textbf{Cross-Dataset Bias.} While datasets are expected to resemble the probability distribution of the real world, the data collection procedure can be biased by human and systematic factors,
	leading to a distribution mismatch between two datasets. Thus, the knowledge transfer from one dataset (e.g., ImageNet) to other new datasets (e.g., GZSL benchmarks) is limited, which results in the extraction of poor-quality visual features on the new datasets. This is known as cross-dataset bias \cite{2011Unbiased}. Domain adaptation techniques can be used to reduce this bias \cite{Khosla2012UndoingTD,Oquab2014LearningAT}, but they are not applicable to GZSL, as the features of GZSL benchmarks are extracted from a pre-trained CNN backbone before sequential learning is conducted. Additionally, although fine-tuning is a typical method to alleviate this problem, it inevitably results in other severer issues, including inefficiency and overfitting \cite{Hendrycks2019UsingPC,Li2020RethinkingTH}. Thus, we attempt to essentially circumvent cross-dataset bias and improve \textit{semantic$\rightarrow$visual} mapping in a unified network using feature refinement for GZSL classification. 
	
	\section{Method}\label{sec3}
	
	\noindent\textbf{Motivation.} As shown in Fig. \ref{fig:problem-definition}, the cross-dataset bias limits knowledge transfer from ImageNet to GZSL benchmarks, resulting in poor-quality visual features being extracted from the GZSL benchmarks (e.g., CUB \cite{Welinder2010CaltechUCSDB2}) by a pre-trained CNN backbone. 
	This hampers the \textit{semantic$\rightarrow$visual} learning, feature synthesis, and GZSL classification. 
	As a result, the upper bound of the recognition performance of GZSL on both seen and unseen classes is potentially limited.
	Although fine-tuning may alleviate this issue to a certain degree, it inevitably results in other severer problems \cite{Hendrycks2019UsingPC,Raghu2019TransfusionUT,Li2020RethinkingTH}. For example, it is difficult to fine-tune on a new small dataset, and it is easy for the model to overfit to seen classes, which ultimately is not conducive to the generalization of GZSL.
	
	These observations prompt us to speculate that the current poor performance of GZSL is closely related to the cross-dataset bias. The experimental results of fine-tuning in \cite{Xian2019FVAEGAND2AF,Narayan2020LatentEF} further support this claim. In other words, we believe that, by alleviating the cross-dataset bias, the visual features of GZSL benchmarks will be enhanced, enabling us to also further improve the GZSL classification. 
	To this end, we propose a novel method, termed feature refinement for generalized zero-shot learning (FREE). 
	Our strategy is to utilize class label supervision and a semantic cycle-consistency constraint to guide the proposed feature refinement (FR) module to learn class- and semantically-relevant feature representations in a unified network. FR can effectively refine the visual features and avoids the inefficiency and overfitting risks of fine-tuning.

	\noindent\textbf{Overview.} The pipeline of FREE is shown in Fig. \ref{fig:FREE}. FREE includes a feature generating VAEGAN (f-VAEGAN) \cite{Xian2019FVAEGAND2AF}, a feature refinement module (FR), and a classifier. In Stage 1, f-VAEGAN aims to learn the \textit{semantic$\rightarrow$visual} mapping ($G$) for visual feature synthesis, while FR tries to learn discriminative representations for the real/synthesized seen visual features. The two are jointly optimized. To learn discriminative features, FR is optimized using our SAMC-loss and a semantic cycle-consistency loss. 
	In Stage 2, we refine the visual features of both seen and unseen classes using the trained FR. The refined real seen visual features and refined synthesized unseen visual features are then fed into a classifier (e.g., softmax) for training, while the refined real unseen visual features are used for testing. Thus, FREE is an \textit{inductive} method.
	
	\noindent\textbf{Notation.} We denote seen data as $\mathcal{S}=\left\{\left({x}_{i}, y_{i}\right)\right\}_{i=1}^{M}$, where ${x}_{i}$ is a visual feature, ${y}_{i}$ is its class label in $\mathcal{Y}^{s}$, and $M$ is the number of seen images.  Let $\mathcal{Y}^{u}$ be the set of unseen classes, which is disjoint from the seen class set $\mathcal{Y}^{s}$, i.e., $\mathcal{Y}^{s} \cap \mathcal{Y}^{u}= \varnothing$. Each seen class and unseen class have their own corresponding attribute embedding ${a}_{j} \in \mathcal{A}, \forall j \in \mathcal{Y}^{s} \cup \mathcal{Y}^{u}$. 
	
	\subsection{Revisiting f-VAEGAN}\label{sec3.1}
	f-VAEGAN \cite{Xian2019FVAEGAND2AF} has achieved impressive performance and thus become a popular baseline for generative GZSL methods. In this paper, we take f-VAEGAN as a baseline for learning \textit{semantic$\rightarrow$visual} mapping. f-VAEGAN integrates a VAE \cite{Kingma2014AutoEncodingVB} and GAN \cite{Goodfellow2014GenerativeAN} into a unified generative model to simultaneously take advantage of both. Specifically, it comprises a feature generating VAE (f-VAE) and a feature generating WGAN (f-WGAN). The f-VAE consists of an encoder $E(x,a)$ (denoted as $E$) and a decoder $G$ (shared with the f-WGAN,  as a conditional generator $G(z,a)$). The encoder $E(x,a)$ encodes an input seen visual feature $x$ to a latent code $z$, while the decoder $G(z,a)$ reconstructs visual feature  $\hat{x}$ from $z$. the f-VAE is first optimized by a VAE loss $\mathcal{L}_{V}$:
	\begin{gather}
	\label{Lv}
	\begin{aligned}
	\mathcal{L}_{V} &= \mathcal{L}_{KL}+ \mathcal{L}_{R\_x}\\
	&=\operatorname{KL}(E(x, a) \| p(z \mid a))-\mathbb{E}_{E(x, a)}[\log G(z, a)],
	\end{aligned}
	\end{gather}
	where $\mathcal{L}_{KL}$ is the Kullback-Leibler divergence,  $p(z \mid a)$ is a prior distribution assumed to be $\mathcal{N}(0,1)$, and $\mathcal{L}_{R\_x}$ is the visual feature reconstruction loss represented by $-\log G(z, a)$. The f-WGAN, on the other hand, comprises a generator $G(z, a)$ and a discriminator $D(x, a)$ (denoted as $D$). The generator $G(z, a)$ synthesizes a visual feature $\hat{x}$ from a random input noise $z$, whereas the discriminator $D(x, a)$ takes a real visual feature $x$ or a synthesized visual feature $\hat{x}$ and outputs a real value indicating the degree of realness or fakeness. Both $G$ and $D$ are conditioned on the embedding $a$, optimized by the WGAN loss
	\begin{gather}
	\label{L_wgan}
	\begin{aligned}
	\mathcal{L}_{W} &=\mathbb{E}[D(x, a)]-\mathbb{E}[D(\hat{x}, a)]\\
	&-\lambda \mathbb{E}\left[\left(\|\nabla D({x^{\prime}}, a)\|_{2}-1\right)^{2}\right],\end{aligned}
	\end{gather}
	where $x^{\prime}=\tau x + (1-\tau) \hat{x}$ with $\tau \sim U(0,1)$ and $\lambda$ is the penalty coefficient.
	
	\subsection{Feature Refinement}\label{sec3.2}
	Our strategy towards circumventing the cross-dataset bias is to refine the visual features of GZSL benchmarks for remedying the limited knowledge transfer with an FR module, which is constrained by the SAMC-loss and semantic cycle-consistency loss. Furthermore, we concatenate the features of various layers in FR to extract the fully refined features for classification.
	
	\noindent\textbf{Self-Adaptive Margin Center Loss.} To encourage FR to learn more discriminative class-relevant representations for visual features, we propose the self-adaptive margin center loss (SAMC-loss, $\mathcal{L}_{S A M C}$) to constraint FR.  There are four reasons of why and how we conduct SAMC-loss: (1) Since class label information is available, we introduce $\mathcal{L}_{SAMC}$ to explicitly encourage intra-class compactness and inter-class separability, and guide FR to learn discriminative class-relevant features. (2) $\mathcal{L}_{SAMC}$ has the advantages of the center loss \cite{Wen2016ADF} and triplet loss \cite{Schroff2015FaceNetAU}, e.g., avoid artificial sampling, and simultaneously learn intra-class compactness and inter-class separability. (3) Considering the intra-class compactness and inter-class separability are differently sensitive to various datasets, i.e., coarse-grained and fine-grained datasets, $\mathcal{L}_{SAMC}$ takes a balance factor ($\gamma$) to balance the inter-class separability and intra-class compactness adaptively. (4) $\mathcal{L}_{SAMC}$ is conducted on the intermediate encoded features $\mu$ in FR, which is directly beneficial for improving the discriminability of features of the shallower layers in FR. Furthermore, the class centers in $\mathcal{L}_{SAMC}$ are dynamically updated during training, enabling the discriminative feature learning to be more effective. $\mathcal{L}_{S A M C}$ is formulated as:
	\begin{align}
	\label{L_smc}
	\mathcal{L}_{S A M C}\left(\hat{a}, y, y^{\prime}\right)=\max \left(0, \Delta+\gamma\left\|\mu-\mathbf{c}_{y}\right\|_{2}^{2}
	\notag\right.
	\\
	\phantom{=\;\;}
	\left.-(1-\gamma)\left\|\mu-\mathbf{c}_{y^{\prime}}\right\|_{2}^{2}\right),
	\end{align}
	where $\mathbf{c}_{y}$ is the $y$th (the label of seen visual feature $x$) class center of semantic embedding, $\mathbf{c}_{y^{\prime}}$ is the $y^{\prime}$th (a randomly selected class label other than $y$) class center, $\Delta$ represents the margin that controls the distance between intra- and inter-class pairs, $\mu$ is the encoded feature in FR, and $\gamma \in [0,1]$ is used for balancing the inter-class separability and intra-class compactness, which are adaptable to various datasets.

	We use a large $\gamma$ for  fine-grained datasets (e.g., CUB, SUN), and a small $\gamma$ for coarse-grained datasets (e.g., AWA1 \cite{Lampert2014AttributeBasedCF}, AWA2 \cite{Xian2019ZeroShotLC}). This is done because (1) when the classes are ambiguous in a fine-grained dataset, we can more easily distinguish them by encouraging intra-class compactness, as shown in Fig. \ref{fig:intuitive-analysis}{\color{red}(a)}; (2) when the classes are confused in a coarse-grained dataset, we can effectively seperate them by enlarging the inter-class separability, as shown in  Fig. \ref{fig:intuitive-analysis}{\color{red}(b)}. Note that our SAMC-loss is different from the max-margin loss \cite{Akata2016LabelEmbeddingFI,Akata2015EvaluationOO,Reed2016LearningDR,Frome2013DeViSEAD}. Our SAMC-loss enables the model to learn visual-semantic embedding and refined features simultaneously. In contrast, the max-margin loss is typically employed to learn a compatibility function \cite{Akata2016LabelEmbeddingFI,Akata2015EvaluationOO,Reed2016LearningDR} or a scoring function \cite{Frome2013DeViSEAD} between images and textual side-information to represent visual-semantic label embedding in conventional ZSL, resulting in limited performance caused by the cross-dataset bias.

	\noindent\textbf{Semantic Cycle-Consistency Loss.} The last layer of FR reconstructs the semantic embedding $\hat{a}$ from  $\hat{x}$ or  $x$ using the reparametrization trick \cite{Kingma2014AutoEncodingVB}. To further guide FR to effectively learn semantically-relevant representations, we apply a semantic cycle-consistency loss ($\mathcal{L}_{R\_a}$) \cite{Felix2018MultimodalCG,Narayan2020LatentEF} to the reconstructed semantic embeddings to ensure that the synthesized semantic vectors $\hat{a}$ are transformed to be the same embeddings that generated them. To this end, semantically-relevant features are learned using FR. The semantic cycle-consistency loss $\mathcal{L}_{R\_a}$ is achieved using the $\ell_{1}$ reconstruction loss, formulated as follows:
	\begin{gather}
	\label{L_Ra}
	\mathcal{L}_{R\_a}=\mathbb{E}\left[\|\hat{a}_{real}-a\|_{1}\right]+\mathbb{E}\left[\|\hat{a}_{syn}-a\|_{1}\right],
	\end{gather}
	where $\hat{a}_{real}$ are the semantically-relevant features synthesized from $x$ using FR, and  $\hat{a}_{syn}$ are those synthesized from $\hat{x}$. Note that $\hat{a}=\hat{a}_{real} \cup \hat{a}_{syn}$, and $a$ is the semantic embedding corresponding to visual features $x$ or $\hat{x}$.

	\begin{figure}[t]
		\begin{center}
			\includegraphics[width=1\linewidth]{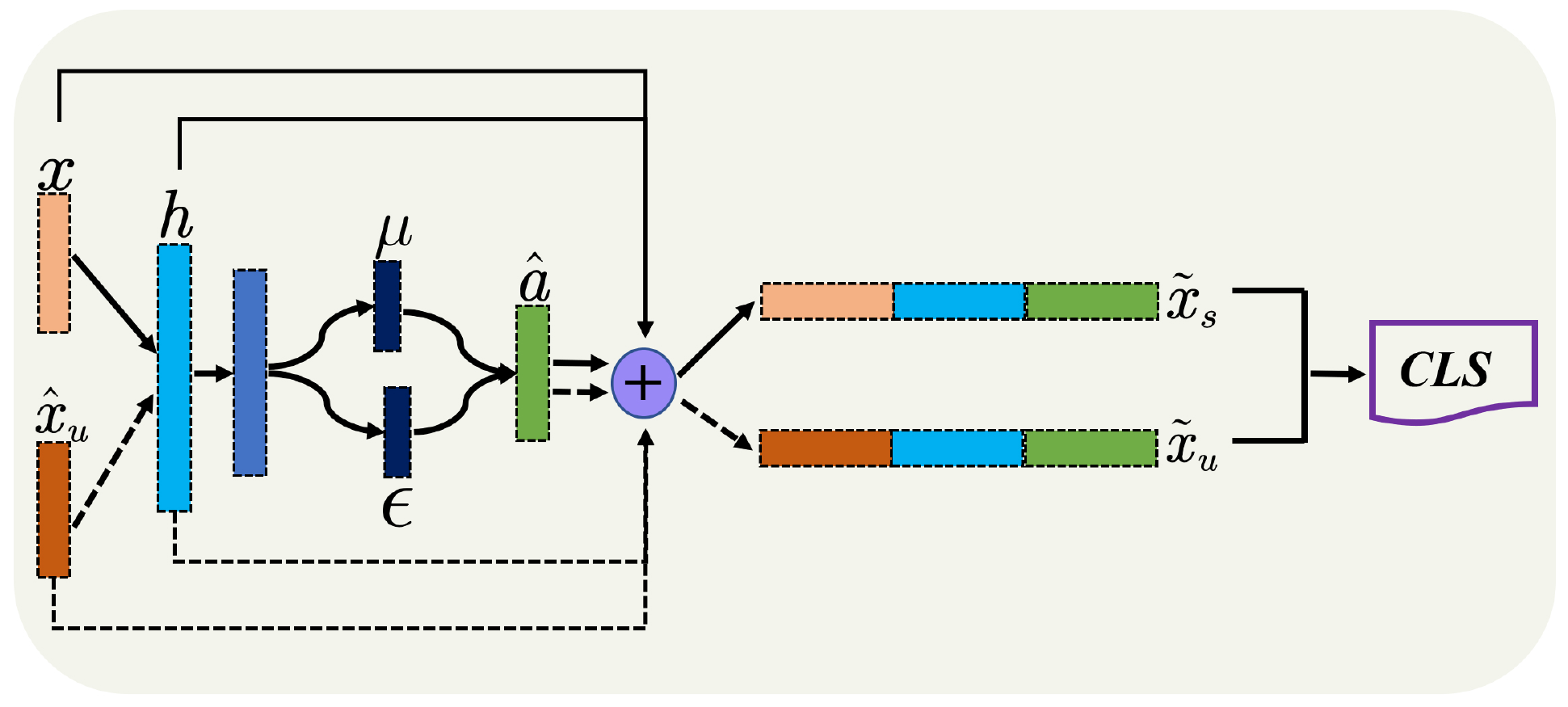}
			
			\caption{Extracting fully refined features in FR. Best viewed in color.}
			\label{fig:feature_refinement}
		\end{center}
	\end{figure}
	
	\noindent\textbf{Extracting Fully Refined Features.} After training in Stage 1, we extract the fully refined features $\tilde{x}_{s}$ and $\tilde{x}_{u}$ from FR to refine real seen visual features $x$ and real/synthesized unseen visual features $x_u/\hat{x}_u$ into discriminative features. FR transforms the original high-dimensional features into low-dimensional features, inevitably discarding some discriminative information, which may hamper the GZSL classification performance.  Using the residual information \cite{He2016DeepRL}, we concatenate the visual features $x$ and $\hat{x}$ with the corresponding latent embedding $h_{s}, h_{u} \in \mathcal{H}$ and semantically-relevant embedding $\hat{a}_s, \hat{a}_u\in \mathcal{A}$ learned from FR as fully refined features, as shown in Fig. \ref{fig:feature_refinement}. They are formulated as:
	\begin{align}
	\label{concatenation}
	\tilde{x}_{s}&= x \oplus h_{s} \oplus \hat{a}_s\\
	\tilde{x}_{u}&= \hat{x}_u \oplus h_{u} \oplus \hat{a}_u
	\end{align}
	where $\oplus$ is a \textit{concatentation} operation, and $\tilde{x}_{s}$ and $\tilde{x}_{u}$ $\in \mathcal{\tilde{X}}$. Thus, the visual features are fully refined as discriminative features, which are class- and semantically-relevant for reducing ambiguities among feature instances of different categories. Note that the refined real seen features and refined synthetic unseen features are used to train the classifier, while the refined real unseen features are used for testing. 
	
	\subsection{Optimization}\label{sec3.3}
	We jointly train the encoder ($E$), generator ($G$), discriminator ($D$), and feature refinement (FR) to optimize the overall objective, which involves a weighted sum of the following losses:
	\begin{gather}
	\label{L_total}
	\begin{aligned}
	\mathcal{L}_{total}(E,G,D,FR) &=\mathcal{L}_V+\mathcal{L}_W\\
	&+\lambda_{SAMC}\mathcal{L}_{S A M C}+\lambda_{R\_a}\mathcal{L}_{R\_a},\end{aligned}
	\end{gather}
	where $\lambda_{SAMC}$ and $\lambda_{R\_a}$ are weights that control the importance of the related loss terms. Similar to the alternative updating policy for GANs, we alternately train $E$, $G$ before the generated visual features and $E$, $D$ and FR after the generated visual features. This is a joint learning framework that couples \textit{semantic$\rightarrow$visual} mapping and visual feature refinement in a unified network. There is also an online interaction between them that mutually benefits both tasks for GZSL. However, fine-tuning is limited in this.

	\subsection{Classification}\label{sec3.4}
	
	\begin{table*}[ht]
		\centering
		\caption{State-of-the-art comparison on five datasets. The best and second-best results are marked in {\color{red}\textbf{red}} and {\color{blue}\textbf{blue}}, respectively. 
		} \label{Table:SOTA}
		\resizebox{!}{4.2cm}
		{
			\begin{tabular}{c|l|ccc|ccc|ccc|ccc|ccc}
				\toprule
				&\multirow{2}*{Method} &\multicolumn{3}{c|}{AWA1} &\multicolumn{3}{c|}{AWA2}&\multicolumn{3}{c|}{CUB}&\multicolumn{3}{c|}{SUN}&\multicolumn{3}{c}{FLO}\\
				&&${U}$&${S}$&${H}$&${U}$&${S}$&${H}$&${U}$&${S}$&${H}$&${U}$&${S}$&${H}$&${U}$&${S}$&${H}$\\
				\midrule
				\multirow{5}*{
					\begin{tabular}{c}
						\rotatebox{90}{\hspace{-2.5cm}Non-generative}
					\end{tabular}
				}
				&DCN~\cite{Liu2018GeneralizedZL}&25.5&\textbf{\color{blue}84.2}&39.1&--&--&--&28.4&60.7&38.7&25.5&37.0&30.2&--&--&--\\
				&SP-AEN~\cite{Chen2018ZeroShotVR}&23.3&{\color{red}\textbf{90.9}}&37.1&--&--&--&34.7&70.6&46.6&24.9&38.6&30.3&--&--&--\\
				&AREN~\cite{Xie2019AttentiveRE}&--&--&--&15.6&\textbf{\color{red}92.9}&26.7&38.9&{\color{red}\textbf{78.7}}&52.1&19.0&\textbf{\color{blue}38.8}&25.5&--&--&--\\
				&CRnet~\cite{Zhang2019CoRepresentationNF}&58.1&74.7&65.4&--&--&--&45.5&56.8&50.5&34.1&36.5&35.3&--&--&--\\
				&GAFE~\cite{Liu2019GraphAA}&25.5&76.6&38.2&26.8&78.3&40.0&22.5&52.1&31.4&19.6&31.9&24.3&--&--&--\\
				&PQZSL~\cite{Li2019CompressingUI}&--&--&--&31.7&70.9&43.8&43.2&51.4&46.9&35.1&35.3&35.2&--&--&--\\
				&MLSE~\cite{Ding2019MarginalizedLS}&--&--&--&23.8&83.2&37.0&22.3&\textbf{\color{blue}71.6}&34.0&20.7&36.4&26.4&--&--&--\\
				&TCN~\cite{Jiang2019TransferableCN}&49.4&76.5&60.0&\textbf{\color{blue}61.2}&65.8&63.4&52.6&52.0&52.3&31.2&37.3&34.0&--&--&--\\
				&DVBE~\cite{Min2020DomainAwareVB}&--&--&--&\textbf{\color{red}63.6}&70.8&\textbf{\color{blue}67.0}&53.2&60.2&\textbf{\color{blue}56.5}&45.0&37.2&40.7&--&--&--\\
				
				\midrule
				\multirow{11}*{
					\begin{tabular}{c}
						\rotatebox{90}{\hspace{-1cm}Generative}
					\end{tabular}
				}
				&SE-GZSL~\cite{Arora2018GeneralizedZL}&56.3&67.8&61.5&58.3&68.1&62.8&41.5&53.3&46.7&40.9&30.5&34.9&--&--&--\\
				&f-CLSWGAN~\cite{Xian2018FeatureGN}&57.9&61.4&59.6&--&--&--&43.7&57.7&49.7&42.6&36.6&39.4&59.0&73.8&65.6\\
				&cycle-CLSWGAN~\cite{Felix2018MultimodalCG}&56.9&64.0&60.2&--&--&--&45.7&61.0&52.3&\textbf{\color{blue}49.4}&33.6&40.0&59.2&72.5&65.1\\
				&CADA-VAE~\cite{Schnfeld2019GeneralizedZA}&57.3&72.8&64.1&55.8&75.0&63.9&51.6&53.5&52.4&47.2&35.7&40.6&--&--&--\\
				&f-VAEGAN~\cite{Xian2019FVAEGAND2AF}&--&--&--&57.6&70.6&63.5&48.4&60.1&53.6&45.1&38.0&41.3&56.8&74.9&64.6\\
				&LisGAN~\cite{Li2019LeveragingTI}&52.6&76.3&62.3&--&--&--&46.5&57.9&51.6&42.9&37.8&40.2&57.7&{\color{blue}\textbf{83.8}}&68.3 \\
				&GMN~\cite{Sariyildiz2019GradientMG}&\textbf{\color{blue}61.1}&71.3&\textbf{\color{blue}65.8}&--&--&--&\textbf{\color{blue}56.1}&54.3&55.2&{\color{red}\textbf{53.2}}&33.0&40.7&--&--&--\\
				&E-PGN~\cite{Yu2020EpisodeBasedPG}&--&--&--&52.6&\textbf{\color{blue}83.5}&64.6&52.0&61.1&56.2&--&--&--&{\color{red}\textbf{71.5}}&82.2&{\color{red}\textbf{76.5}}\\
				&OCD-CVAE~\cite{Keshari2020GeneralizedZL}&--&--&--&59.5&73.4&65.7&44.8&59.9&51.3&44.8&{\color{red}\textbf{42.9}}&\textbf{\color{red}43.8}&--&--&--\\
				&LsrGAN~\cite{Vyas2020LeveragingSA}&54.6&74.6&63.0&--&--&--&48.1&59.1&53.0&44.8&37.7&40.9&--&--&--\\
				
				\cline{2-17}
				&&&&&&&&&&&&&&&&\\
				&\textbf{FREE (Ours)}&\textbf{\color{red}62.9}&69.4&	\textbf{\color{red}66.0}&60.4& 75.4&\textbf{\color{red}67.1}&\textbf{\color{red}55.7} &59.9&\textbf{\color{red}57.7}&47.4&37.2&\textbf{\color{blue}41.7}&\textbf{\color{blue}67.4}&\textbf{\color{red}84.5}&\textbf{\color{blue}75.0}\\
				\bottomrule
			\end{tabular}
		}
	\end{table*}
	
	After obtaining the training data, we train a supervised classifier in the refined feature space as the final GZSL classifier. GZSL aims to learn the classifier  $f_{g z s l}: \mathcal{\tilde{X}} \rightarrow \mathcal{Y}^{s} \cup \mathcal{Y}^{u}$. During testing, the seen/unseen test features are refined as new features by FR, and then used for further testing.
	
	\begin{table}[h]
		\centering
		\caption{Statistics of the CUB, SUN, FLO, AWA1 and AWA2 datasets, including the dimensions of semantic vectors per class (\textit{Att}), seen/unseen class
			size (\textit{Seen/Unseen}), and total number of images (\textit{Img}).} \label{Table:dataset}
		\setlength{\tabcolsep}{4mm}{
			\begin{tabular}{l|c|c|c}
				\hline       
				&  \textit{Att}  &      \textit{Seen/Unseen}       & \textit{Img} \\   
				\hline
				CUB   & 312 &  150/50   & 11788  \\
				SUN   & 102 &  645/72   & 14340  \\
				FLO   & 1024  &  82/20  & 8189   \\
				AWA1  & 85  &  40/10    & 30475  \\
				AWA2  & 85  &  40/10    & 37322  \\
				\hline
				
		\end{tabular}}
	\end{table}
	
	\section{Experiments}\label{sec4}
	\noindent\textbf{Datasets.} We evaluate our method on five benchmark datasets, i.e., CUB (Caltech UCSD Birds 200) \cite{Welinder2010CaltechUCSDB2}, SUN (SUN Attribute) \cite{Patterson2012SUNAD},  FLO (Oxford Flowers) \cite{Nilsback2008AutomatedFC}, AWA1 (Animals with Attributes 1) \cite{Lampert2014AttributeBasedCF}, and  AWA2 (Animals with Attributes 2)\cite{Xian2019ZeroShotLC}. Among these, CUB, SUN and FLO  are fine-grained datasets, whereas AWA1 and AWA2  are coarse-grained. We use the same seen/unseen splits and class embeddings, as \cite{Xian2019ZeroShotLC}, which are summarized in Table \ref{Table:dataset}.

	\noindent\textbf{Evaluation Protocols.} During testing, we use the unified evaluation protocol proposed from \cite{Xian2019ZeroShotLC} to facilitate direct comparison.  Since the test set is composed of seen classes ($\mathcal{Y}^{s}$) and unseen classes ($\mathcal{Y}^{u}$), we evaluate the top-1 accuracies on both, denoted as S and U, respectively. Furthermore, their harmonic mean (defined as $H =(2 \times S \times U) /(S+U)$) is also used for evaluating the performance of GZSL.

	\noindent\textbf{Implementation Details.} In FREE, the encoder, generator and discriminator are multilayer perceptrons (MLPs) containing a 4096-unit hidden layer with LeakyReLU activation. FR is also an MLP that has two hidden layers with 4096-unit and $2\times |\hat{a}|$-unit activated by LeakyReLU, followed an encoding layer where the second hidden layer is encoded into two feature vectors with the size of $|\hat{a}|$. Its output layer $\hat{a}$ is learned by the reparametrization trick \cite{Kingma2014AutoEncodingVB} and corresponds to the semantic vector of the datasets (e.g., $|\hat{a}|=312$ for CUB). We use the Adam optimizer \cite{Kingma2015AdamAM} with $\beta_1$ = 0.5, $\beta_2$ = 0.999. The visual features are extracted from the 2048-dimimentional top-layer pooling units of a ResNet-101 pre-trained on ImageNet, following \cite{Xian2018FeatureGN}. The penalty coefficient $\lambda$ is set to 10. We empirically set the loss weights $\lambda_{SAMC}$ and $\lambda_{R\_a}$ to 0.5 and 0.1/0.001 respectively for SUN/others datasets. The balance factor $\gamma$ is set to 0.8 and 0.1 for fine- and coarse-grained datasets, respectively.

	\begin{figure*}[ht]
		\begin{center}
			
			\includegraphics[width=4.2cm,height=3cm]{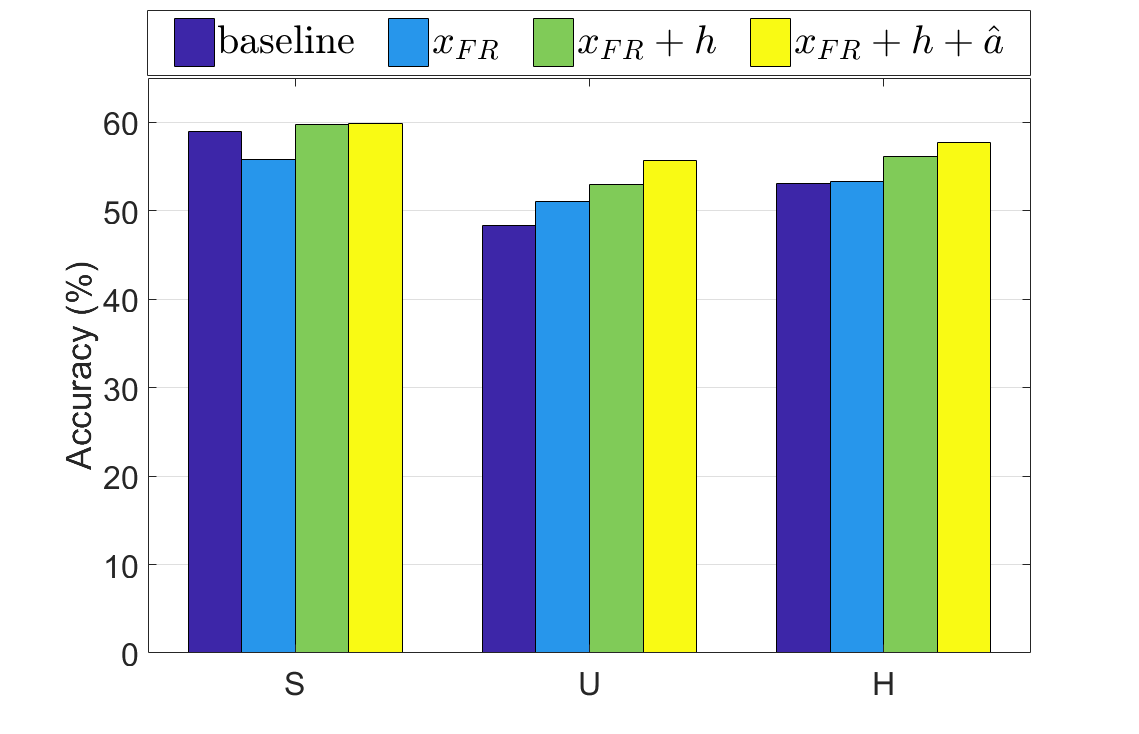}\hspace{1mm}
			\includegraphics[width=4.2cm,height=3cm]{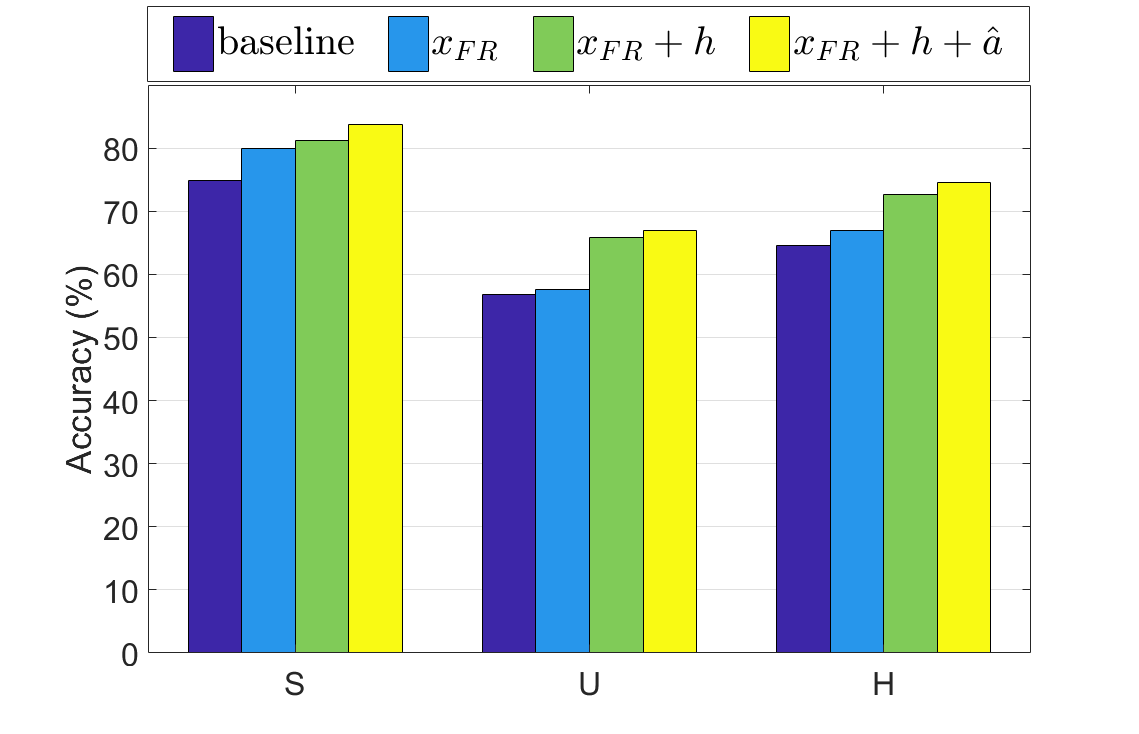}\hspace{1mm}
			\includegraphics[width=4.2cm,height=3cm]{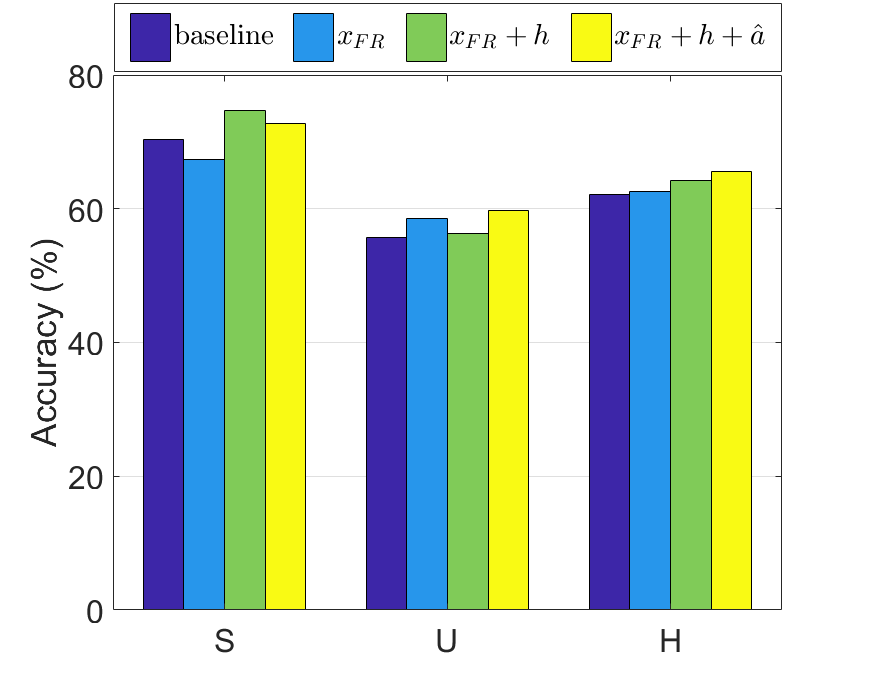}\hspace{1mm}
			\includegraphics[width=4.2cm,height=3cm]{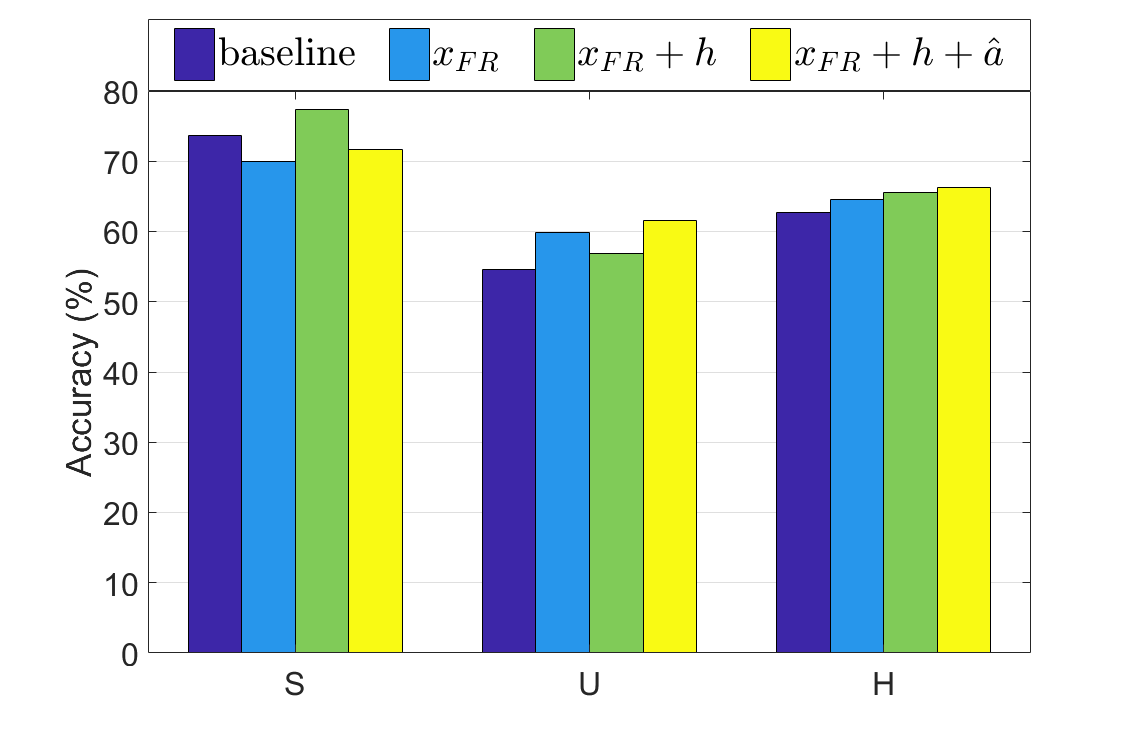}\\
			(a) CUB  \hspace{2.8cm}  (b) FLO \hspace{2.8cm}  (c) AWA1  \hspace{2.4cm}  (d) AWA2 
			\caption{The effectiveness of various visual feature components refined by FR. Best viewed in color.}
			\label{fig:feature-components}
		\end{center}
	\end{figure*}

	\subsection{Comparison with State of the Arts }\label{sec4.1}
	Since FREE is an inductive method, we compare it with other state-of-the-art inductive models for a fair comparison. We categorize the compared methods into generative and non-generative methods. 
	
	Table \ref{Table:SOTA} shows the top-1 accuracies of different methods on unseen classes (U), seen classes (S) and their harmonic mean (H). The results show that FREE consistently attains the best performance for harmonic mean on three benchmarks, i.e., 66.0 on AWA1, 67.1 on AWA2, and 57.7 on CUB. Meanwhile, FREE achieves the second-best results for harmonic mean with 41.7 and 75.0 on SUN and FLO, respectively. These results indicate that the refined features are discriminative and generic for seen/unseen classes on both coarse- and fine-grained datasets. Notably, unlike the compared state-of-the-art methods which only achieve good performance on either seen or unseen classes, FREE attains promising results on both. This reveals that FREE maintains a good balance between seen and unseen classes, benefitting from the unified model jointly trained for \textit{semantic$\rightarrow$visual} mapping and FR. Specifically, the joint training enables the two modules to encode complementary information of categories and encourages them to learn discriminative representations by avoiding cross-dataset bias.
	\begin{figure}[t]
		\begin{center}
			
			\includegraphics[width=4.3cm,height=2.4cm]{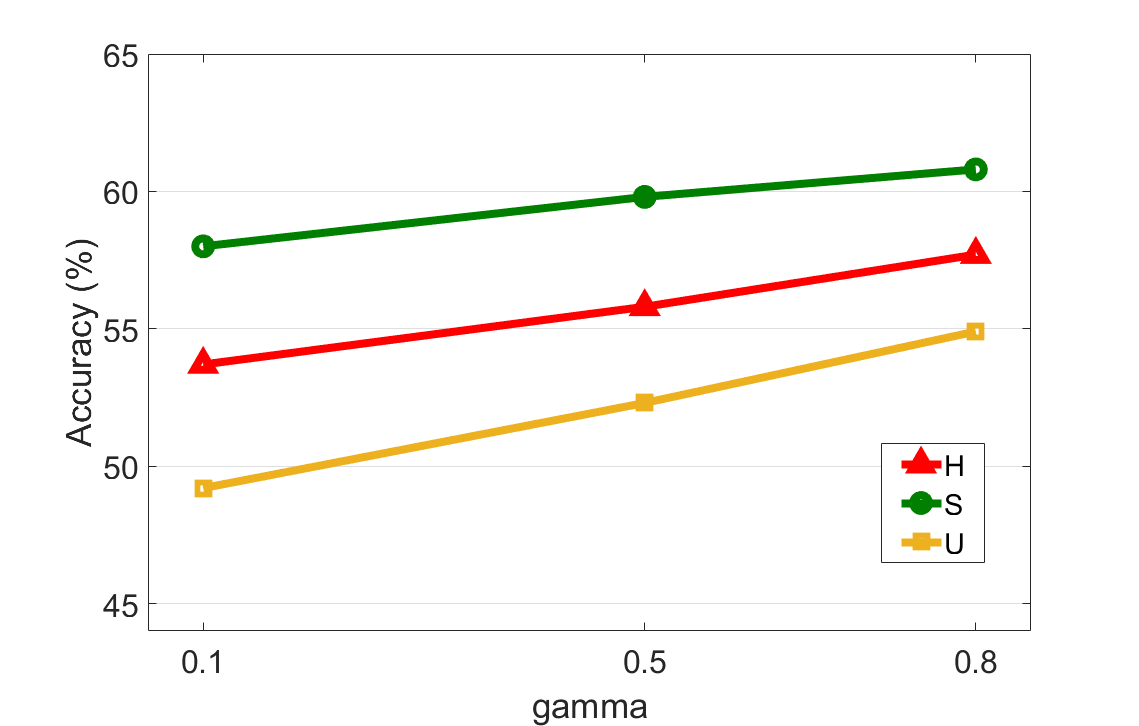}\hspace{-4mm}
			\includegraphics[width=4.3cm,height=2.4cm]{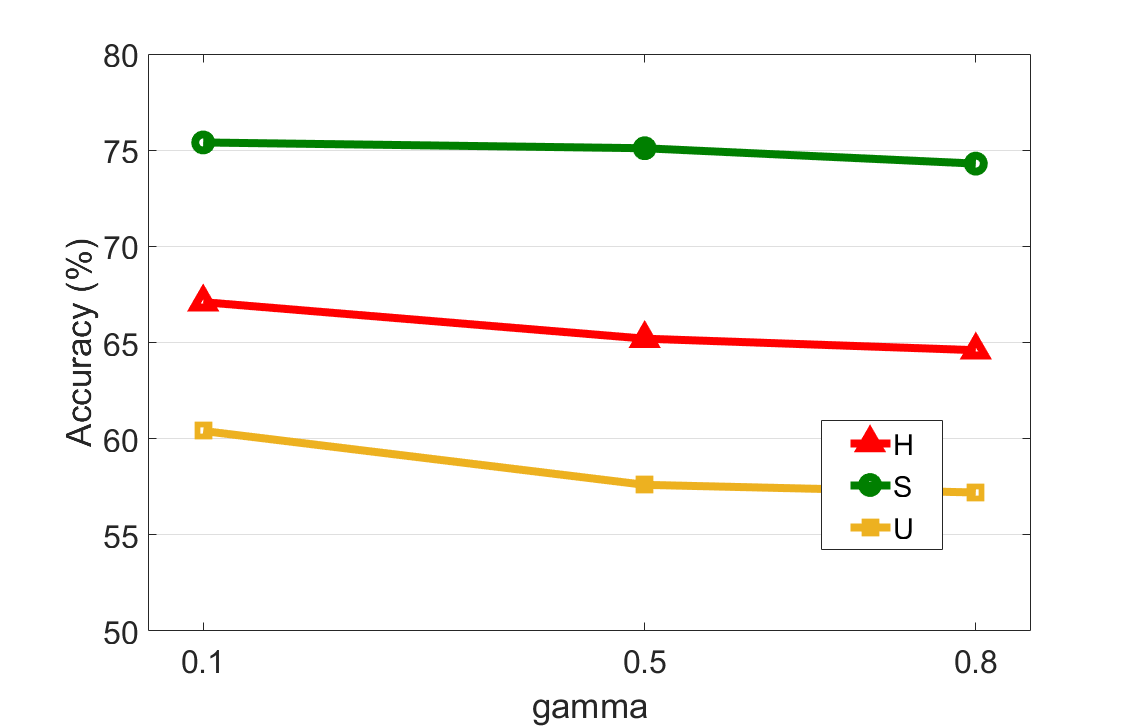}\\
			(a) CUB  \hspace{2.8cm}  (b) AWA2
			\caption{The effectiveness of the balance factor $\gamma$ for the SAMC-loss.}
			\label{fig:gamma}
		\end{center}
	\end{figure}

	\begin{table}[t]
		\centering
		\caption{ Ablation studies for different components of FREE on CUB and  AWA2. The best results are marked in \textbf{boldface}.
		} \label{Table:ablation}
		\resizebox{!}{1.0cm}
		{
			\begin{tabular}{l|ccc|ccc|ccc}
				\toprule
				\multirow{2}*{Method} &\multicolumn{3}{c|}{CUB} &\multicolumn{3}{c|}{FLO} &\multicolumn{3}{c}{AWA2}\\
				&${U}$&${S}$&${H}$&${U}$&${S}$&${H}$&${U}$&${S}$&${H}$\\
				\hline
				f-VAEGAN \cite{Xian2019FVAEGAND2AF}&48.4&60.1&53.6&56.8& 74.9 &64.6 &57.6&70.6&63.5\\
				baseline&48.3&58.9&53.1&57.2& 75.1&64.9&54.6&73.6&62.7\\
				baseline+FR($\mathcal{L}_{R\_a}$)&50.4&\textbf{62.2}&55.7&63.2&73.8&68.1 &57.1&73.8&64.4\\
				baseline+FR($\mathcal{L}_{S A M C}$)&53.2&60.2&56.5&64.0& 82.4&72.0 &59.1&72.7&65.2\\
				baseline+FR($\mathcal{L}_{S A M C}$+$\mathcal{L}_{R\_a}$)&\textbf{54.9}&60.8&\textbf{57.7}&\textbf{67.4}&\textbf{84.5}&\textbf{75.0} &\textbf{60.4}& \textbf{75.4}&\textbf{67.1}\\
				\bottomrule
			\end{tabular}
		}
	\end{table}

	\subsection{Ablation Study}\label{sec4.2}
	To provide further insight into FREE, we conduct ablation studies to evaluate the effect of different model built and feature components. Since FREE is based on f-VAEGAN \cite{Xian2019FVAEGAND2AF}, we re-implement this method as our \textit{baseline}.
	
	\noindent\textbf{Analysis of Model Components.}   As shown in Table \ref{Table:ablation}, our FR provides significant improvements over the baseline with various model components (i.e., FR($\mathcal{L}_{R\_a}$), FR($\mathcal{L}_{S A M C}$) and both combined). We first evaluate the two components independently. FR($\mathcal{L}_{S A M C}$) and FR($\mathcal{L}_{R\_a}$) individually outperform the baseline in harmonic mean on CUB (by 3.4\% and 2.6\%), FLO (by 7.1\% and 3.2\%) and AWA2 (by 2.5\% and 1.7\%). This shows the effectiveness of FR. Interestingly, since the cross-dataset bias on fine-grained datasets (e.g., CUB, FLO) is larger than on coarse-grained datasets (e.g., AWA2), our FR can achieve greater improvements on fine-grained datasets than on coarse-grained datasets. Furthermore, FR($\mathcal{L}_{S A M C}$) performs better than FR($\mathcal{L}_{R\_a}$), which shows the effectiveness of our SAMC-loss. The complete version of FREE gives the highest results on all datasets, achieving an impressive accuracy gain of 4.6\%, 10.1\% and 4.4\% in harmonic mean on CUB, FLO and AWA2, respectively. This indicates that the SAMC-loss and semantic cycle-consistency loss are mutually complementary for FR. These results prove that FR is good for feature enhancement in GZSL, and thus the cross-dataset bias is alleviated.

	\begin{figure}[t]
		\begin{center}
			
			\includegraphics[width=4.3cm,height=2.4cm]{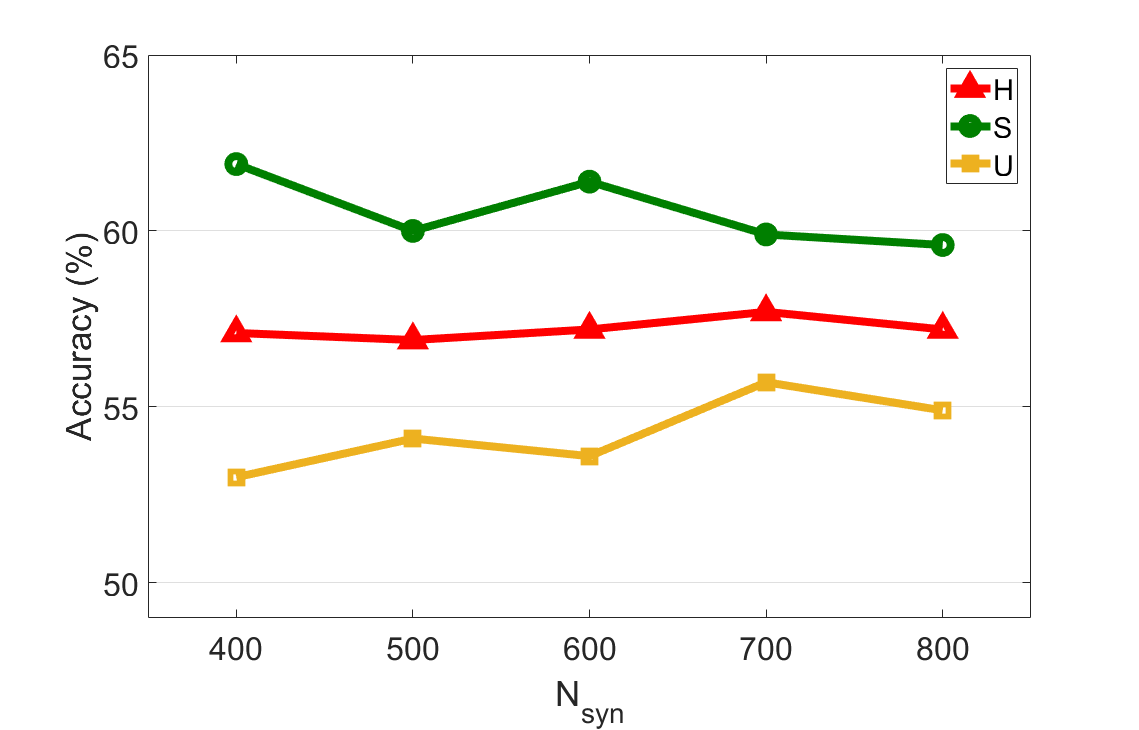}\hspace{-4mm}
			\includegraphics[width=4.3cm,height=2.4cm]{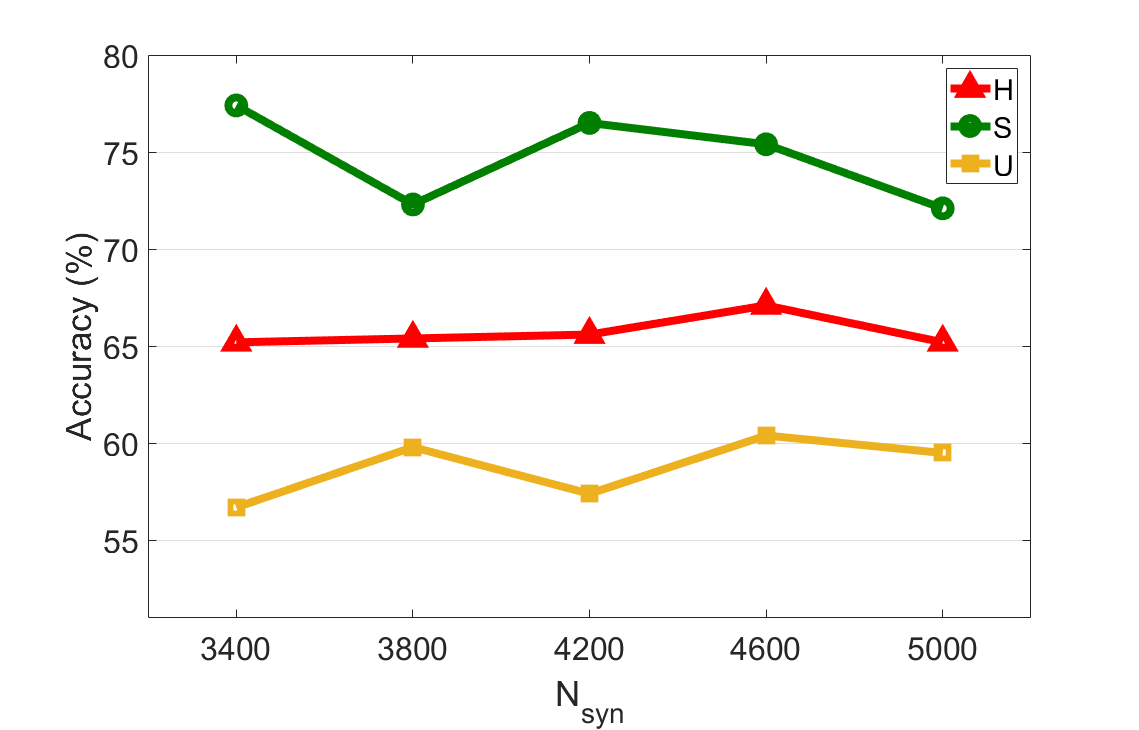}\\
			(a) CUB  \hspace{2.8cm}  (b) AWA2
			\caption{The impact of the number of synthetic visual features $N_{syn}$ in each unseen class.}
			\label{fig:syn_num}
		\end{center}
	\end{figure}

	\begin{figure*}[ht]
		\begin{center}
			\hspace{0.5mm}\rotatebox{90}{\hspace{0.8cm}{\footnotesize (a) CUB }}\hspace{-1mm}
			\includegraphics[width=3cm,height=2.5cm]{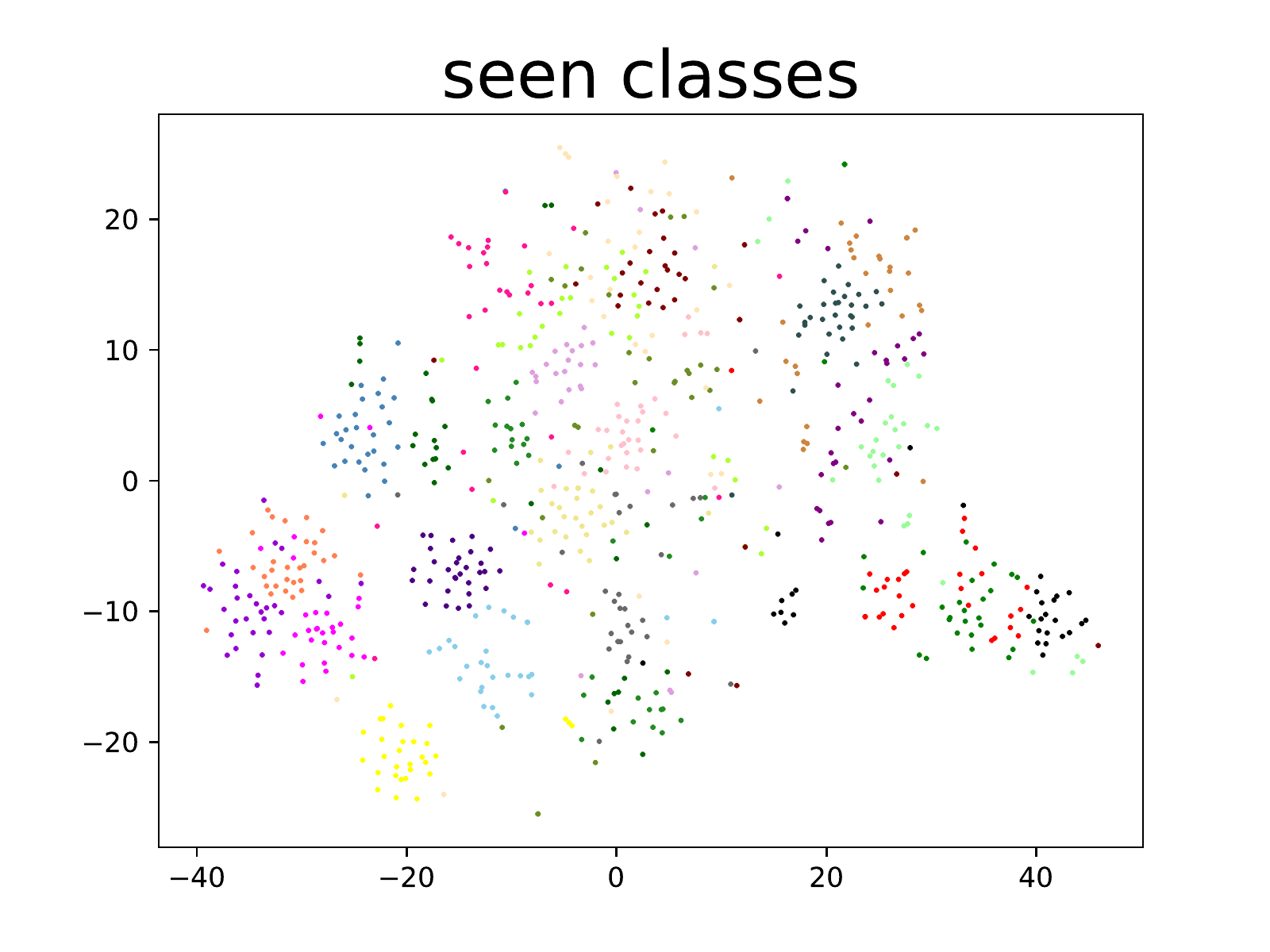}\hspace{-3mm}
			\includegraphics[width=3cm,height=2.5cm]{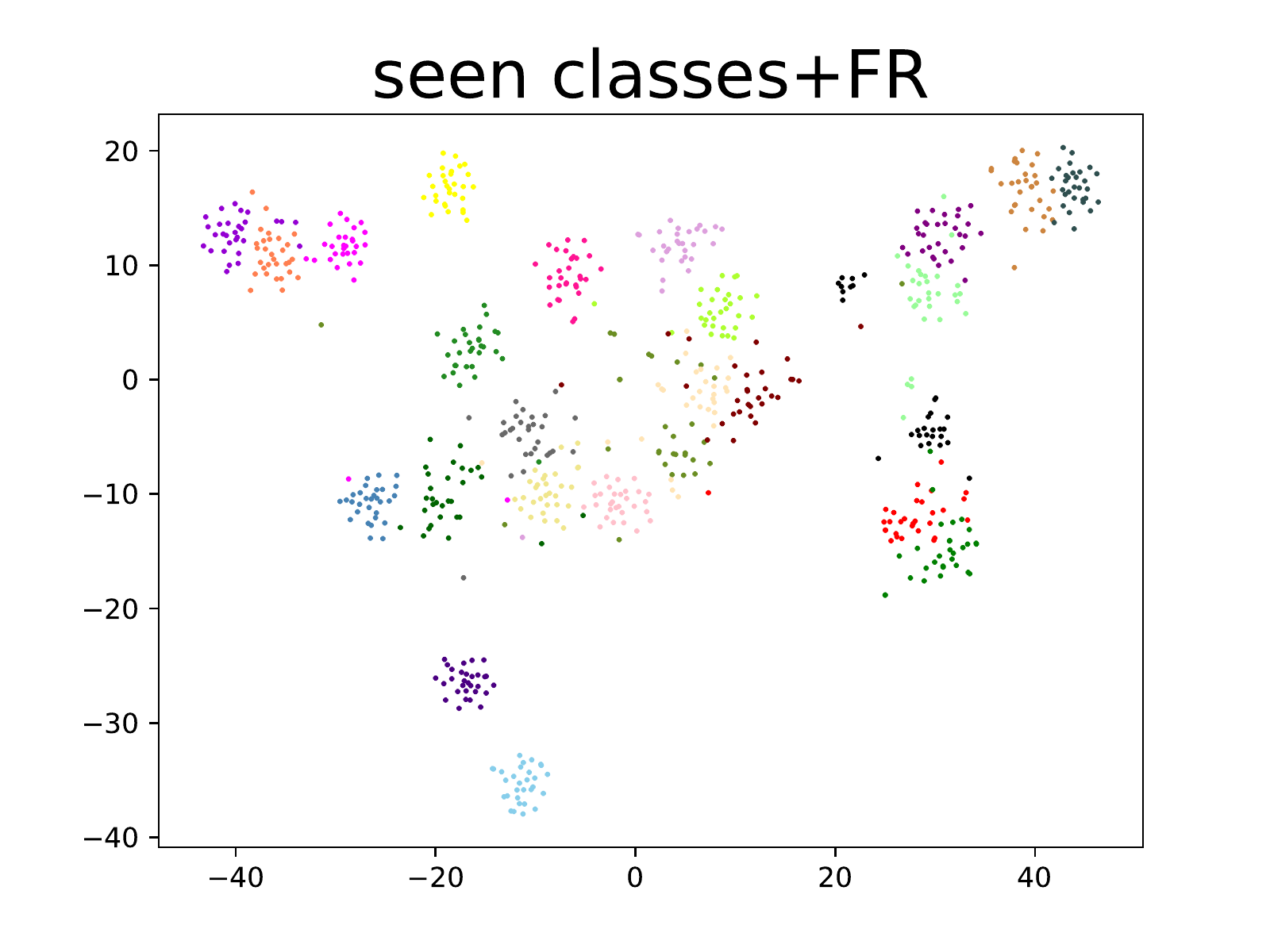}\hspace{-3mm}
			\includegraphics[width=3cm,height=2.5cm]{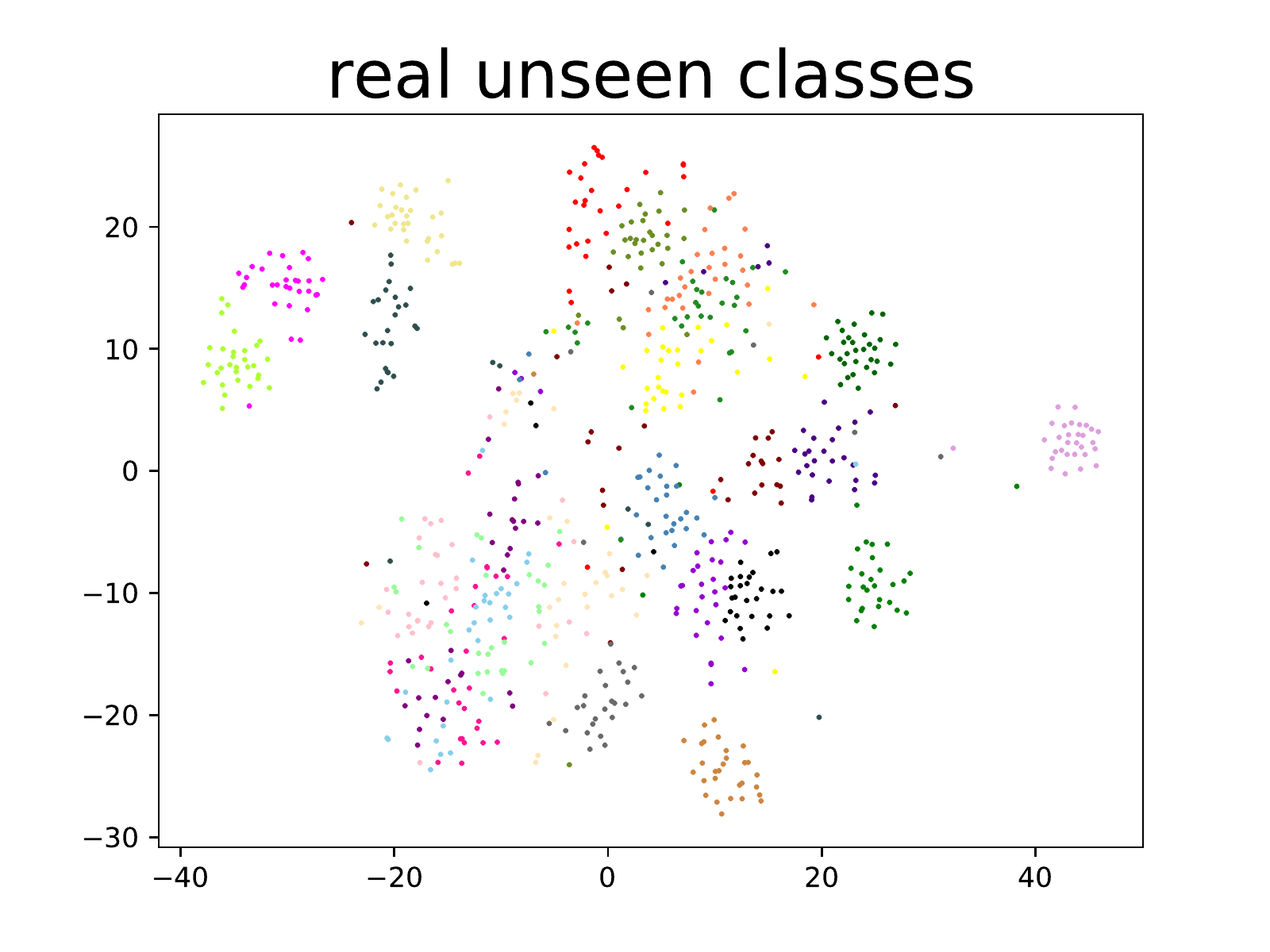}\hspace{-3mm}
			\includegraphics[width=3cm,height=2.5cm]{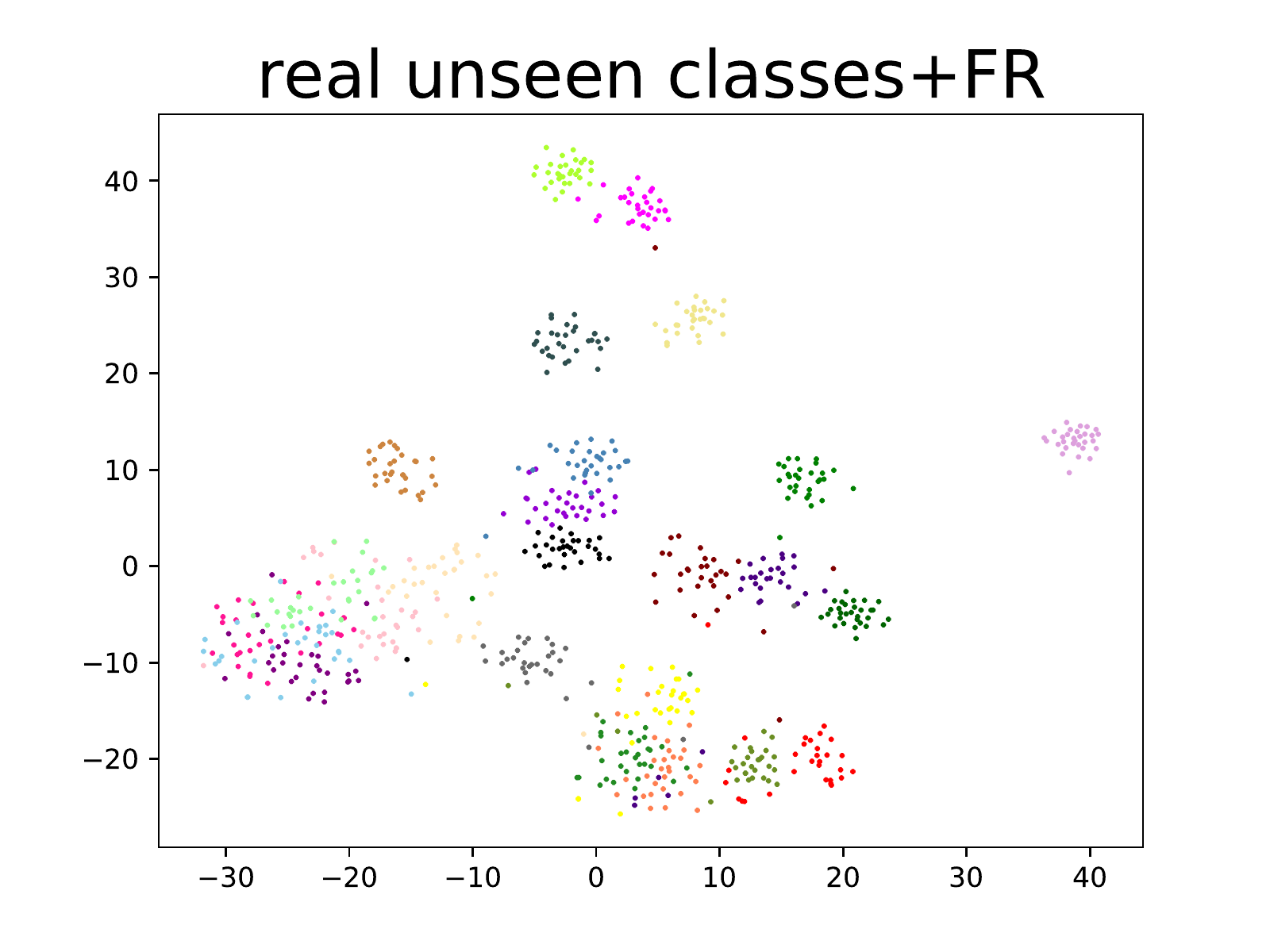}\hspace{-3mm}
			\includegraphics[width=3cm,height=2.5cm]{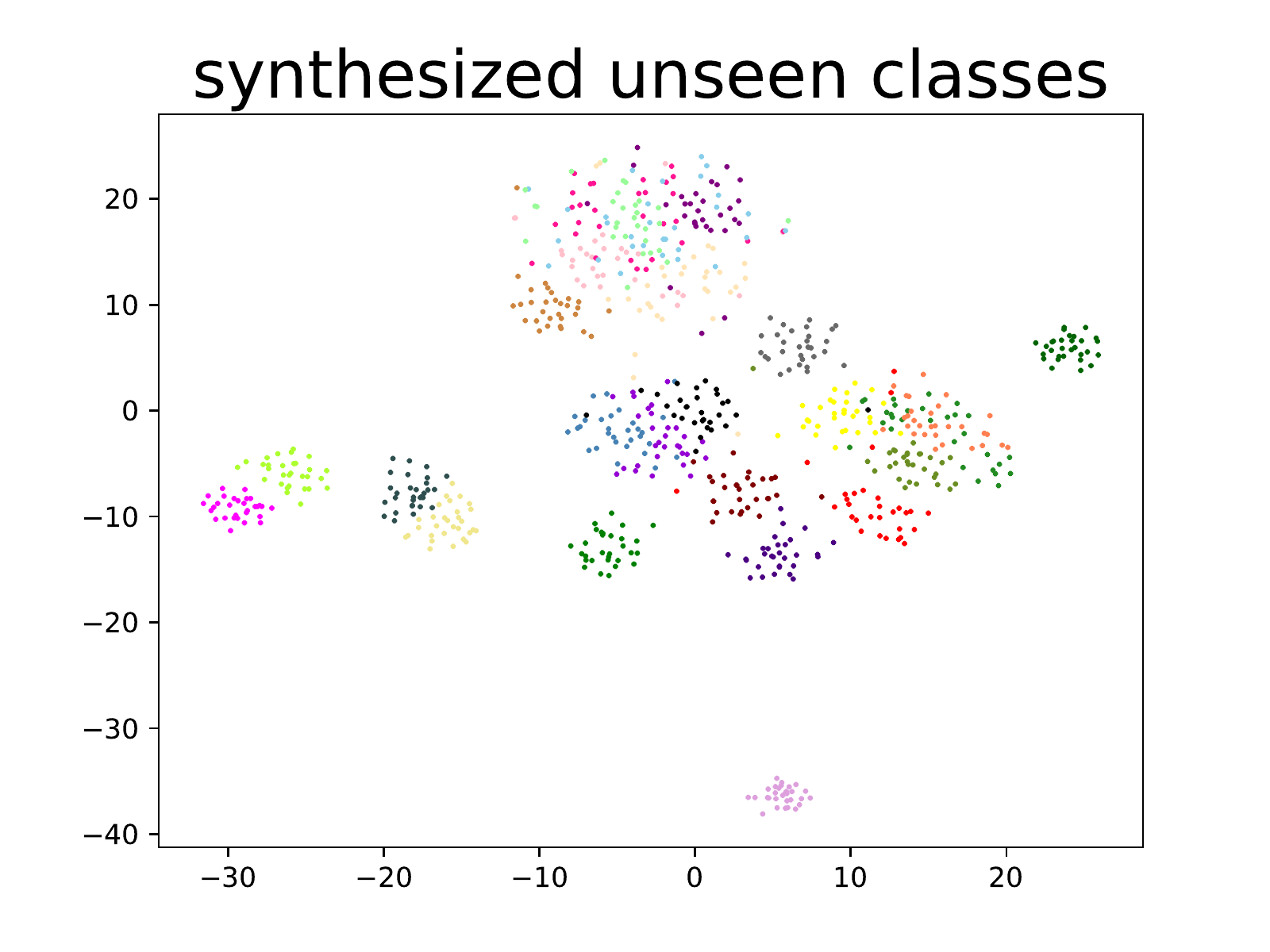}\hspace{-3mm}
			\includegraphics[width=3cm,height=2.5cm]{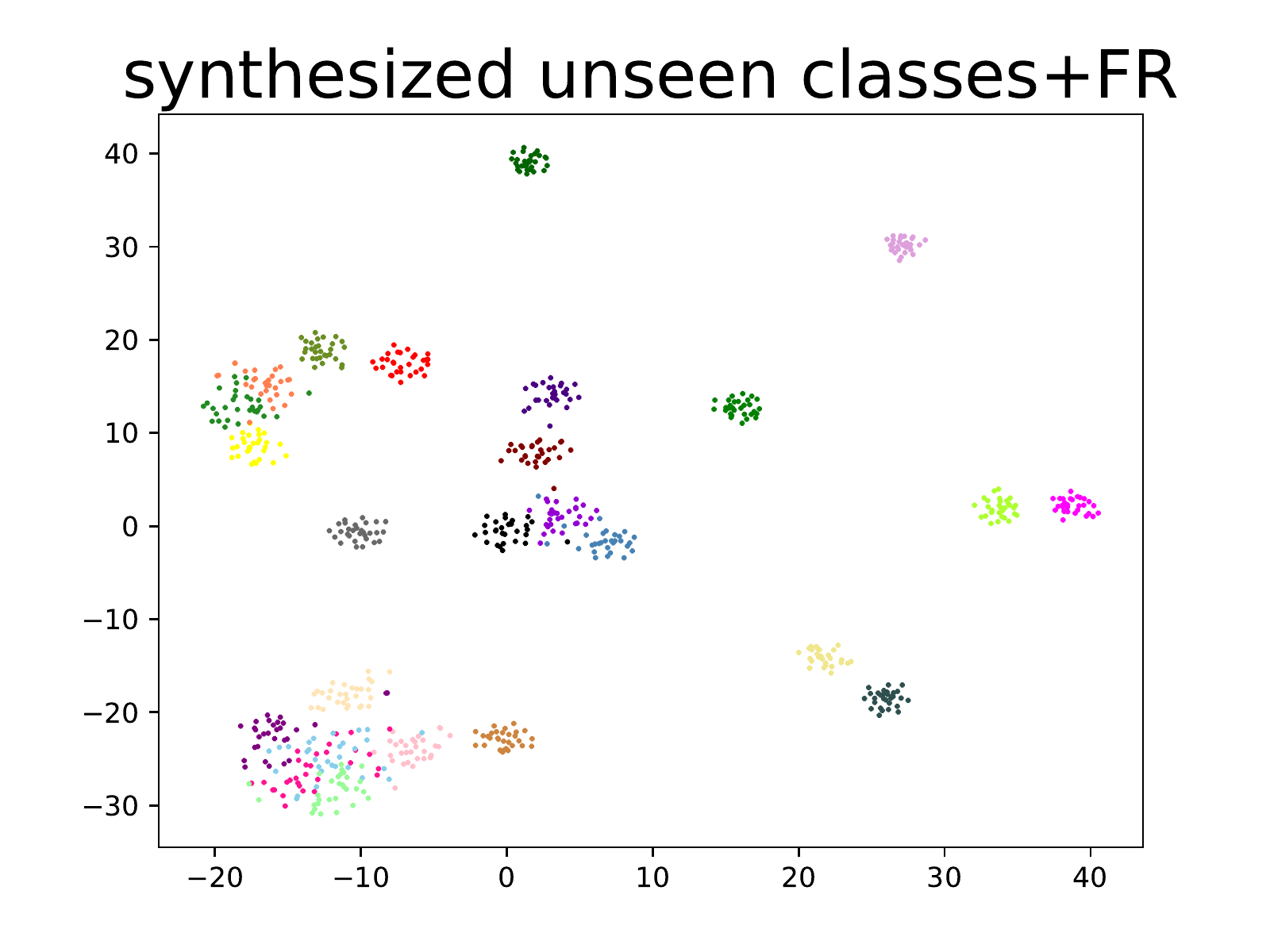}\\
			\vspace{-1.5mm}
			\hspace{0.5mm}\rotatebox{90}{\hspace{0.6cm}{\footnotesize (b) AWA2 }}\hspace{-1mm}
			\includegraphics[width=3cm,height=2.5cm]{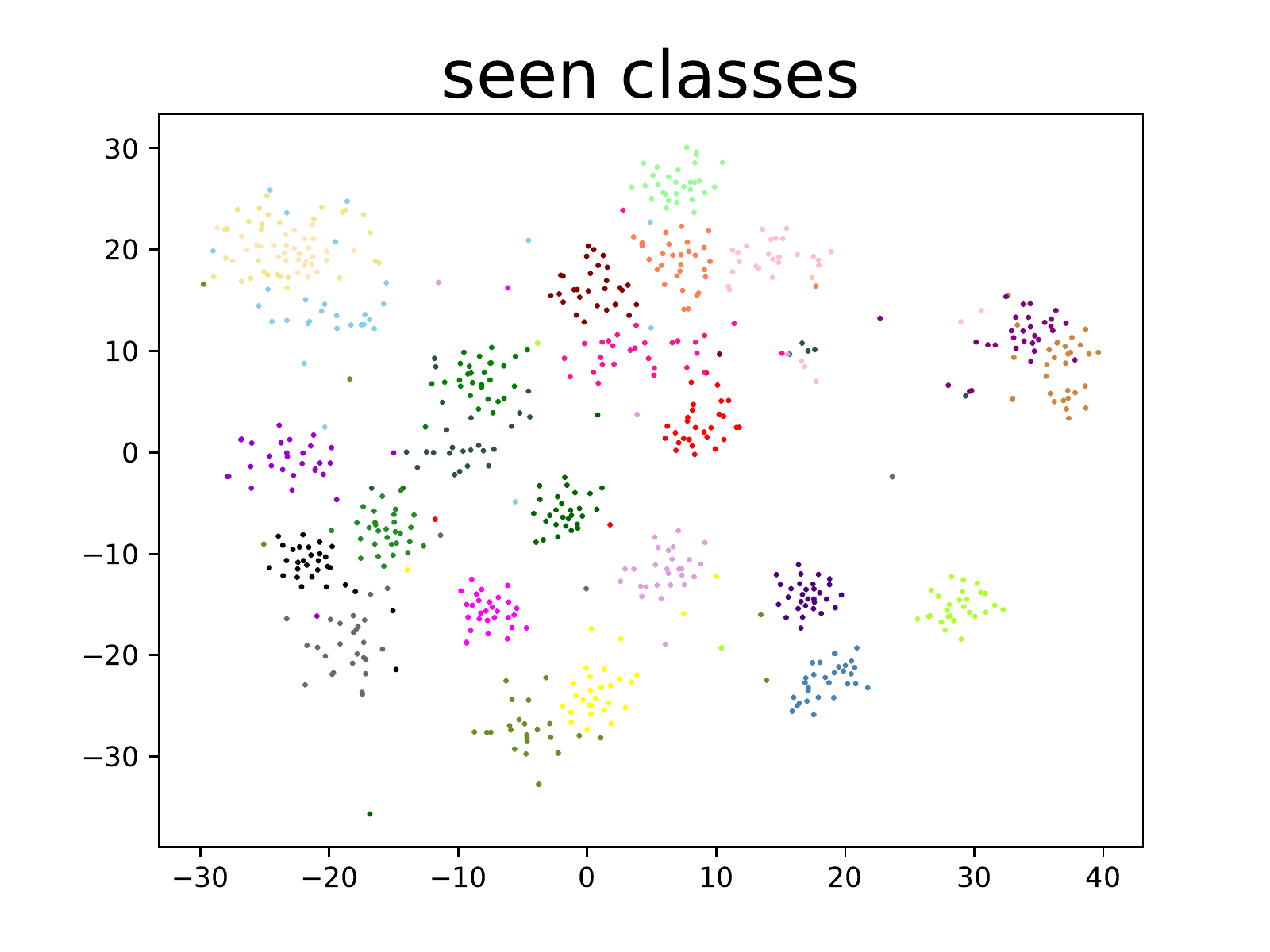}\hspace{-3mm}
			\includegraphics[width=3cm,height=2.5cm]{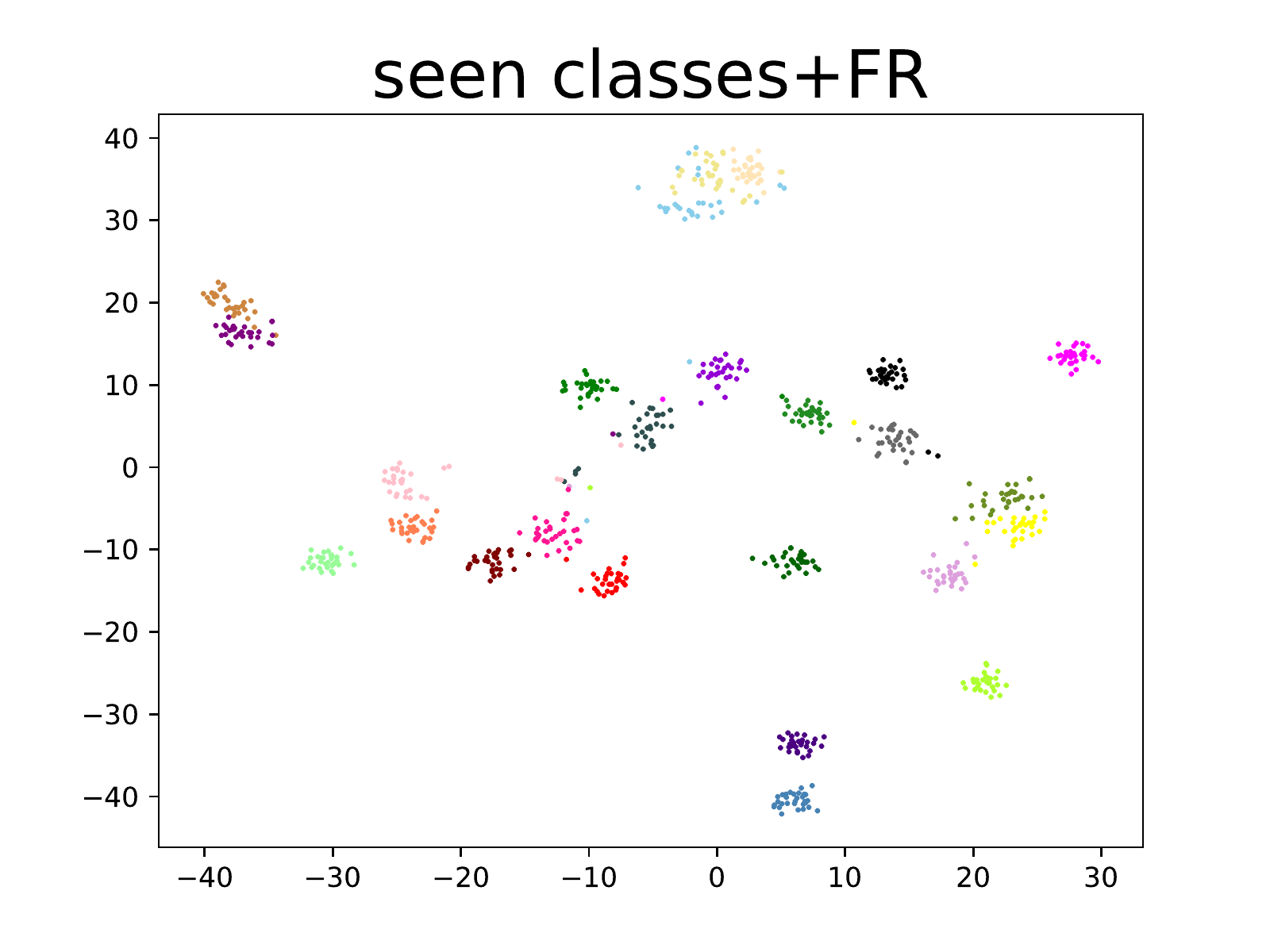}\hspace{-3mm}
			\includegraphics[width=3cm,height=2.5cm]{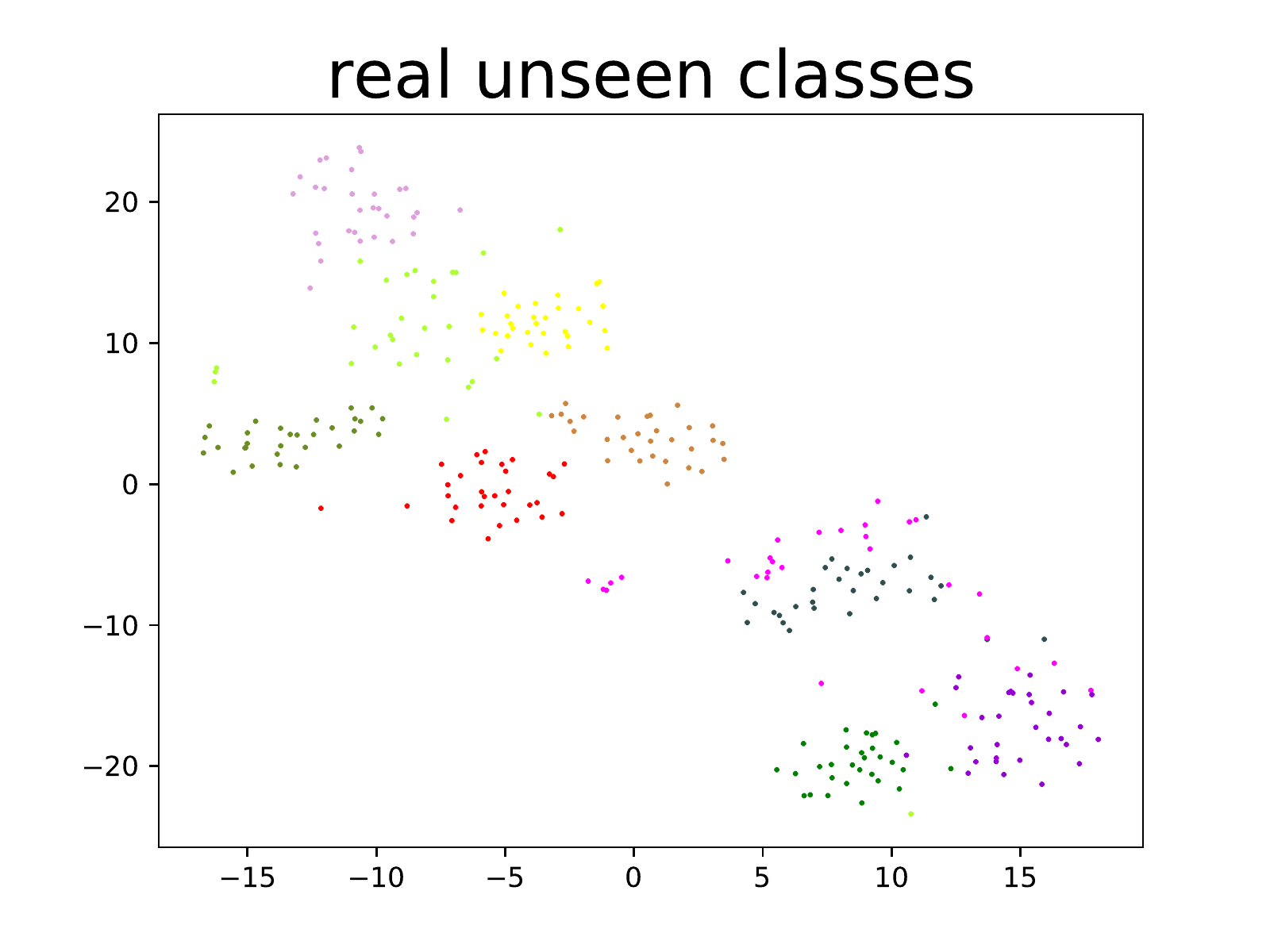}\hspace{-3mm}
			\includegraphics[width=3cm,height=2.5cm]{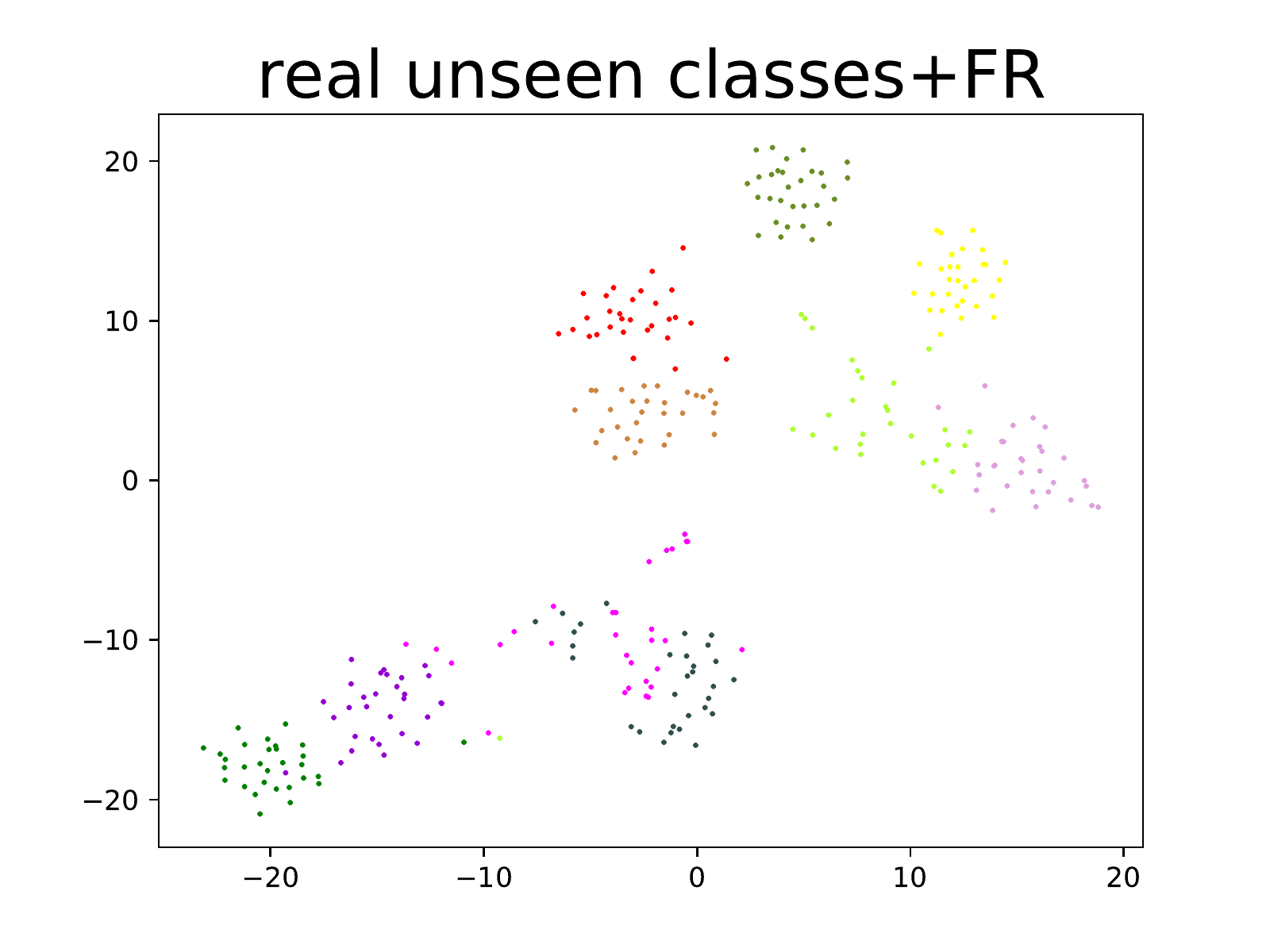}\hspace{-3mm}
			\includegraphics[width=3cm,height=2.5cm]{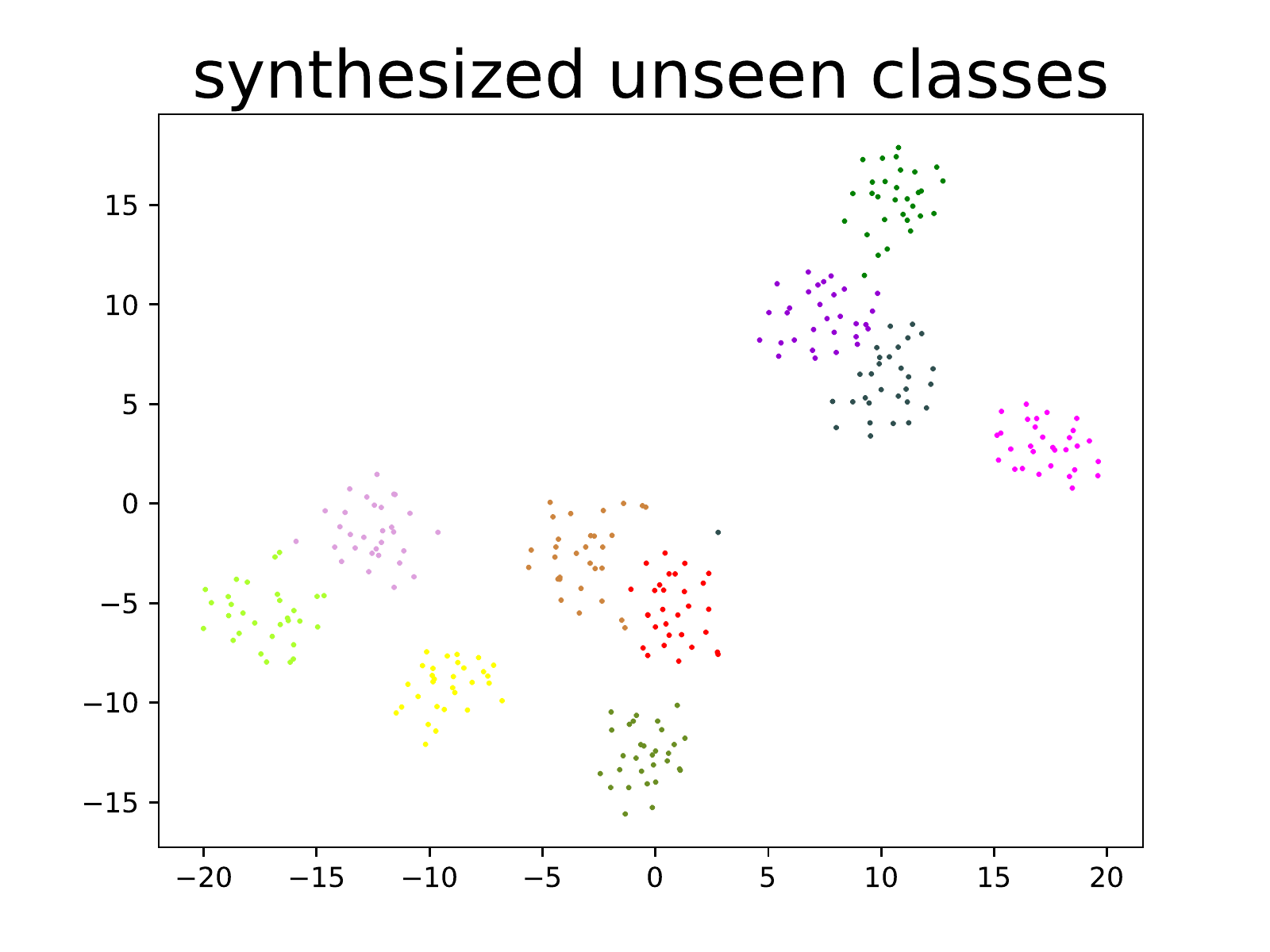}\hspace{-3mm}
			\includegraphics[width=3cm,height=2.5cm]{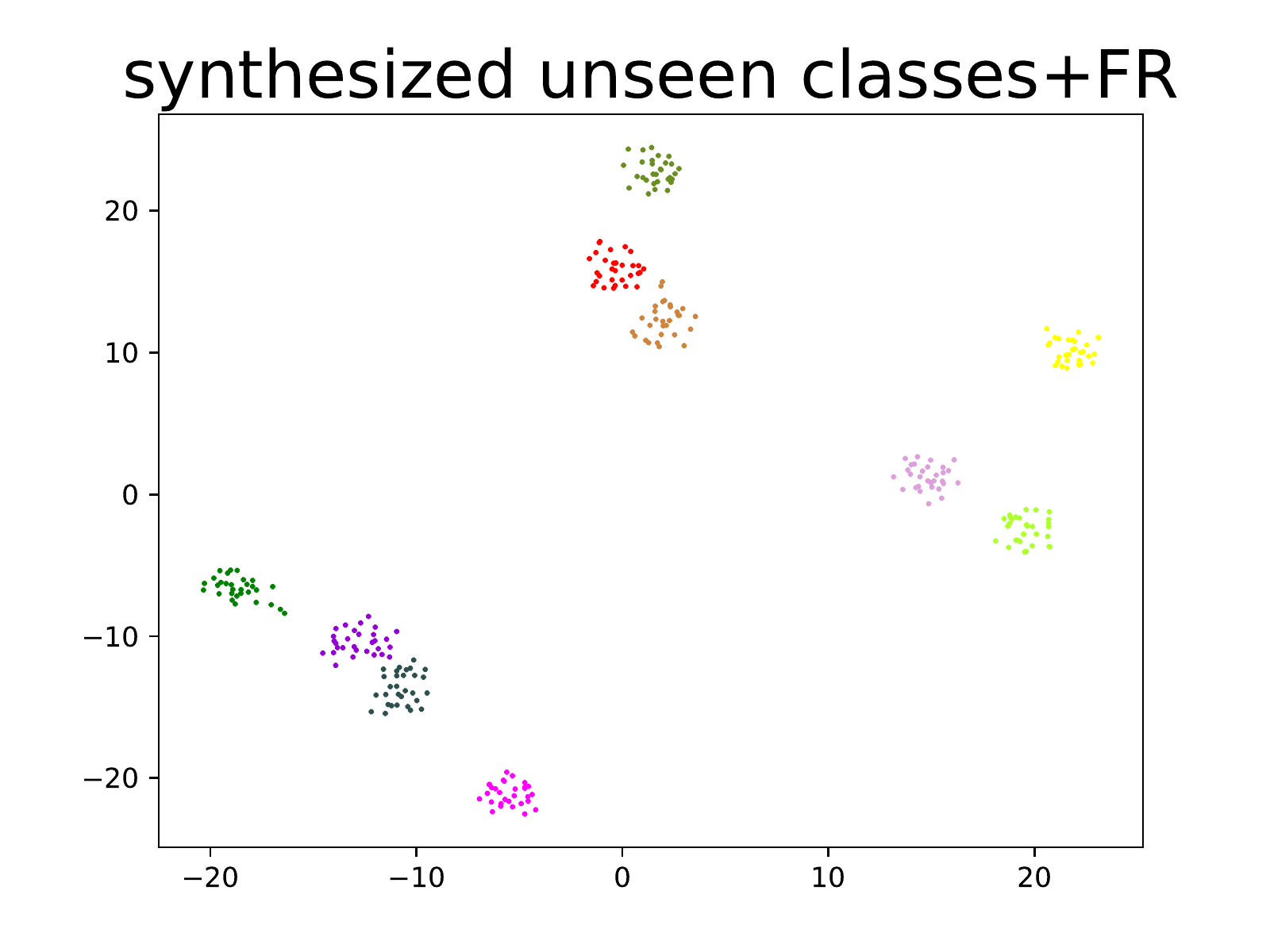}\\\vspace{-1.5mm}
			\caption{t-SNE visualization \cite{Maaten2008VisualizingDU} of visual features for real seen classes, real/synthesized unseen classes, and their features refined by FR on (a) CUB and (b) AWA2. The real/synthesized unseen class features are represented with the same color. Best viewed in color.}
			\label{fig:intuitive-analysis}
		\end{center}
	\end{figure*}
	
	\noindent\textbf{Analysis of Feature Components.} We study the effectiveness of various visual feature components refined by FR. As shown in Fig. \ref{fig:feature-components}, all feature components in FR substantially improve the performance. When only taking the real seen features $x$ and the synthesized unseen visual features $\hat{x}_u$ for classification, labeled as $\bm{x_{FR}}$, FREE provides consistent improvement of harmonic mean over the baseline. This proves that FR contributes positively to the \textit{semantic$\rightarrow$visual} mapping. We also concatenate seen/unseen visual features with the hidden features $h$ (labeled as $\bm{x_{FR}+h}$), and both $h$ and the learned semantically-relevant features $\hat{a}$ (labeled as $\bm{x_{FR}+h+\hat{a}}$)  for classification. Further improvement is achieved. Since there is a more obvious cross-dataset bias for fine-grained datasets (e.g., FLO), FR consistently achieves improvements on all evaluation protocols with various visual feature components. As for coarse-grained datasets (e.g., AWA2) with small cross-dataset bias, FR also improves the performances of unseen accuracy and Harmonic mean. This is attributed to the fact that FR can reduce the cross-dataset bias for improving the quality of visual features.

	
	\subsection{Hyperparameter Analysis}\label{sec4.3}
	
	\noindent\textbf{Balance Factor $\gamma$.} We study the balance factor $\gamma$ in Eq. \ref{L_smc} to determine its influence on the module. As can be seen from Fig. \ref{fig:gamma}, as $\gamma$ grows, S, U and H gain consistent improvement on the fine-grained datasets (e.g., CUB). Nevertheless, S, U and H consistently decrease when $\gamma$ increases on coarse-grained datasets (e.g., AWA2). These results are explained as follows: (1) On the fine-grained datasets, increasing the intra-class compactness provides larger gains when the classes are confused. (2) On the coarse-grained datasets, increasing the inter-class separability significantly benefits the classification of ambiguous classes.

	\noindent\textbf{ Number of Synthesized Visual Features $N_{syn}$.} We evaluate the impact of the number of synthetic visual features per unseen class $N_{syn}$. As shown in Fig. \ref{fig:syn_num}, FREE is generally insensitive to $N_{syn}$ on all datasets. When increasing the number of synthetic features, the seen class accuracy drops slightly and the unseen class accuracy improves. This demonstrates that FREE can also alleviate the seen-unseen bias problem. Since there exists an upper bound on synthetic diversity, all results will decrease if $N_{syn}$ is set to too large. As such, we set $N_{syn}$ to 4600, 4600, 2400, 700 and 300 for AWA1, AWA2, FLO, CUB and SUN, respectively.

	\section{Discussion}\label{sec5}
	\noindent\textbf{FR for Cross-Dataset Bias.} As displayed in Fig. \ref{fig:problem-definition}, we intuitively show that the cross-dataset bias results in poor-quality visual features, which potentially limits recognition performance for both seen and unseen classes of GZSL. The experimental results of fine-tuning in \cite{Xian2019FVAEGAND2AF,Narayan2020LatentEF} well support this claim. In this paper, we attempt to enhance the visual features of seen/unseen classes to address the limited knowledge transfer caused by this bias, using feature refinement (FR), which is encouraged to learn class- and semantically-relevant feature representations. As shown in Fig. \ref{fig:intuitive-analysis}, FR significantly improves the visual features of seen/unseen classes, reducing the ambiguity between different categories. These results intuitively show the effectiveness of FR. Interestingly, the synthesized unseen features share similar class relationships with the real unseen features, which proves that FR helps FREE to learn a promising \textit{semantic$\rightarrow$visual} mapping.  As a result, FREE achieves an impressive performance gain over current state-of-the-art methods and its baseline.

	\noindent\textbf{Feature Refinement vs Fine-tuning.} As analyzed in Section \ref{sec1}, although fine-tuning may alleviate the cross-dataset bias for GZSL to a degree, it leads to ineffectiveness and overfitting (e.g., seen-unseen bias is 29.1\% on FLO shown in Table \ref{Table:fr_fine}). Nevertheless, our proposed FR combines itself with \textit{semantic$\rightarrow$visual} mapping, the two of which are mutually beneficial. Meanwhile, the SAMC-loss and semantic cycle-consistency loss guide FR to learn discriminative feature representations to refine the visual features, and thus the limited knowledge transfer is alleviated and the cross-dataset bias is reduced. As shown in Table \ref{Table:fr_fine}, FR achieves competitive results over fine-tuning on FLO and AWA2, which well supports our claims. 
	
	
	\begin{table}[h]
		\centering
		\caption{ Feature refinement vs fine-tuning on FLO and AWA2. The best results are marked in \textbf{boldface}.
		} \label{Table:fr_fine}
		\resizebox{!}{1.0cm}
		{
			\begin{tabular}{l|ccc|ccc}
				\toprule
				\multirow{2}*{Method}&\multicolumn{3}{c|}{FLO} &\multicolumn{3}{c}{AWA2}\\
				&${U}$&${S}$&${H}$&${U}$&${S}$&${H}$\\
				\hline
				f-VAEGAN \cite{Xian2019FVAEGAND2AF}&56.8&74.9&64.6&57.6&70.1&63.5\\
				f-VAEGAN\textbf{+}finetuned \cite{Xian2019FVAEGAND2AF}&63.3&\textbf{92.4}&\textbf{75.1}&57.1&\textbf{76.1}&65.2\\
				f-VAEGAN\textbf{+}FR&\textbf{67.4}&84.5&75.0&\textbf{60.4}& 75.4&\textbf{67.1}\\
				\bottomrule
			\end{tabular}
		}
	\end{table}
	
	\section{Conclusion}\label{6}
	
	In this paper, we propose a joint learning framework, termed FREE, that couples \textit{semantic$\rightarrow$visual} mapping and FR to alleviate the cross-dataset bias. We further introduce a SAMC-loss that cooperates with the semantic cycle-consistency constraint to encourage FR to learn class- and semantically-relevant feature representations. Meanwhile, we extract the features of various layers in FR as fully refined features for classification. Competitive results on five popular benchmarks demonstrate the superiority and great potential of our approach. Since cross-dataset bias is a major bottleneck for improving the GZSL performance and FR is an effective approach to address it, we believe FR may work well in other generative GZSL methods.
	
	\section*{Acknowledgements} This work is partially supported by NSFC~(61772220) and Key R\&D Plan of Hubei Province~(2020BAB027).
	
	{\small
		\bibliographystyle{ieee_fullname}
		\bibliography{mybibfile}
	}

\end{document}